\documentclass[runningheads,a4paper]{llncs}

\usepackage{amssymb}
\usepackage{graphicx}
\usepackage[caption=false]{subfig}
\usepackage{multirow}
\usepackage{makecell}
\usepackage{xcolor}
\usepackage{siunitx}
\usepackage{breakcites}

\usepackage{url}

\newcommand{\todo}[1]{}
\newcommand{\disable}[1]{}
\renewcommand{\orcidID}[1]{}
\newcommand\rna[1]{{#1}}
\newcommand\rnb[1]{{#1}}
\newcommand\rnc[1]{{#1}}
\newcommand\rnd[1]{{#1}}

\usepackage{hyperref}
\title{\LARGE System for multi-robotic exploration of underground environments\\CTU-CRAS-NORLAB in the DARPA Subterranean Challenge 
	\thanks{The presented work has been supported by the Czech Science Foundation (GAČR) under reseach projects No. 18-18858S, 19-20238S, 20-10280S, 20-27034J, and 20-29531S, and by OP VVV MEYS RCI project CZ.02.1.01/0.0/0.0/16\_019/0000765.}%
}

\author{
	 Tom{\'a}{\v s} Rou{\v c}ek\inst{1}\orcidID{0000-0003-3598-1630} \and
	 Martin Pecka\inst{1}\orcidID{0000-0002-0815-304X} \and
	 Petr {\v C}{\'i}{\v z}ek\inst{1}\orcidID{0000-0001-6722-3928} \and
	 Tom{\'a}{\v s} Pet{\v r}{\'i}{\v c}ek\inst{1}\orcidID{0000-0002-3136-5673} \and
	 Jan Bayer\inst{1}\orcidID{0000-0002-8898-4512} \and 
	 Vojt{\v e}ch {\v S}alansk{\'y} \inst{1}\orcidID{0000-0003-2977-0976} \and
   Teymur Azayev \inst{1}\orcidID{0000-0003-2267-5722}
	 Daniel He{\v r}t \inst{1}\orcidID{0000-0003-1637-6806} \and
	 Mat{\v e}j Petrl{\' i}k\inst{1}\orcidID{0000-0002-5337-9558} \and
	 Tom{\'a}{\v s} B{\' a}{\v c}a\inst{1}\orcidID{0000-0001-9649-8277} \and
	 Vojt{\v e}ch Spurn{\' y}\inst{1}\orcidID{0000-0002-9019-1634} \and 
	 V{\'i}t Kr{\'a}tk{\'y}\inst{1}\orcidID{0000-0002-1914-742X} \and 
   Pavel Petr{\'a}{\v c}ek\inst{1}\orcidID{0000-0002-0887-9430} \and 
	 Dominic Baril\inst{2}\orcidID{0000-0002-7283-8406} \and
	 Maxime Vaidis\inst{2}\orcidID{0000-0002-1749-7207} \and
	 Vladim{\' i}r Kubelka\inst{2}\orcidID{0000-0001-8393-9969} \and
	 Fran\c{c}ois Pomerleau\inst{2}\orcidID{0000-0003-1288-2744} \and
	 Jan Faigl\inst{1}\orcidID{0000-0002-6193-079} \and
	 Karel Zimmermann\inst{1}\orcidID{0000-0002-8898-4512} \and
	 Martin Saska\inst{1}\orcidID{0000-0001-7106-3816} \and
	 Tom{\'a}{\v s}  Svoboda\inst{1}\orcidID{0000-0002-7184-1785} \and 
	 Tom{\'a}{\v s} Krajn{\'i}k\inst{1}\orcidID{0000-0002-4408-7916}    
}

\authorrunning{Rou{\v c}ek et al.}
\titlerunning{DARPA SubT Challenge: Underground multi-robot exploration system}
\institute{Faculty of Electrical Engineering, Czech Technical University \url{roucek.tomas@fel.cvut.cz}\and
  Universit\'{e} Laval, Canada
}

\begin{document}
\maketitle

\begin{abstract}
    We present a field report of CTU-CRAS-NORLAB team from the Subterranean Challenge (SubT) organised by the Defense Advanced Research Projects Agency (DARPA). 
    The contest seeks to advance technologies that would improve the safety and efficiency of search-and-rescue operations in GPS-denied environments. 
    During the contest rounds, teams of mobile robots have to find specific objects while operating in environments with limited radio communication, e.g. mining tunnels, underground stations or natural caverns.
    We present a heterogeneous exploration robotic system of the CTU-CRAS-NORLAB team, which achieved the third rank at the SubT Tunnel and Urban Circuit rounds and surpassed the performance of all other non-DARPA-funded teams. 
    The field report describes the team's hardware, sensors, algorithms and strategies, and discusses the lessons learned by participating at the DARPA SubT contest.
\end{abstract}
\section{Introduction}\label{sec:introduction}

\begin{figure}[!htb]
\begin{center}
    \includegraphics[height=0.5\columnwidth]{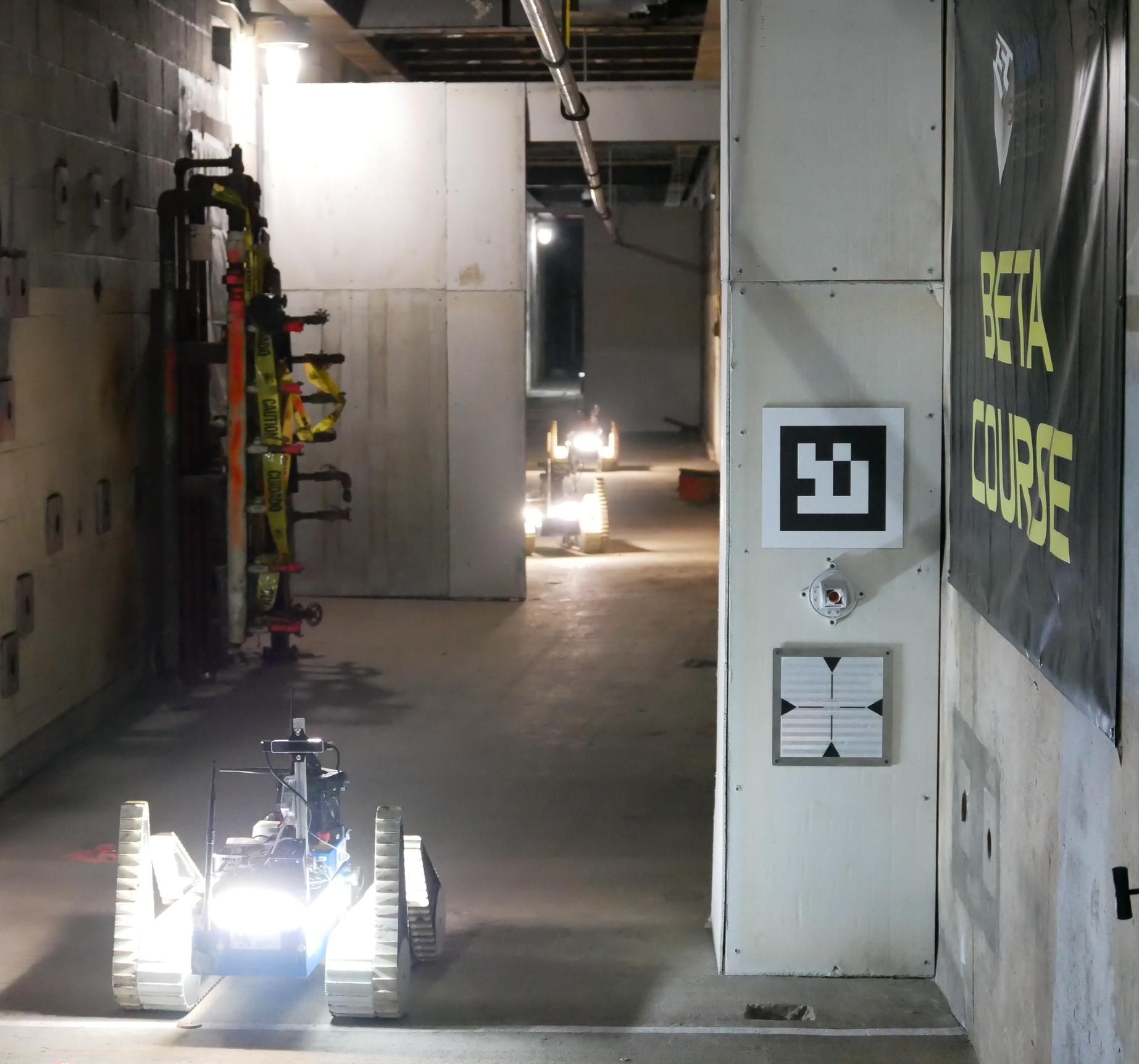}\hfill
    \includegraphics[height=0.5\columnwidth]{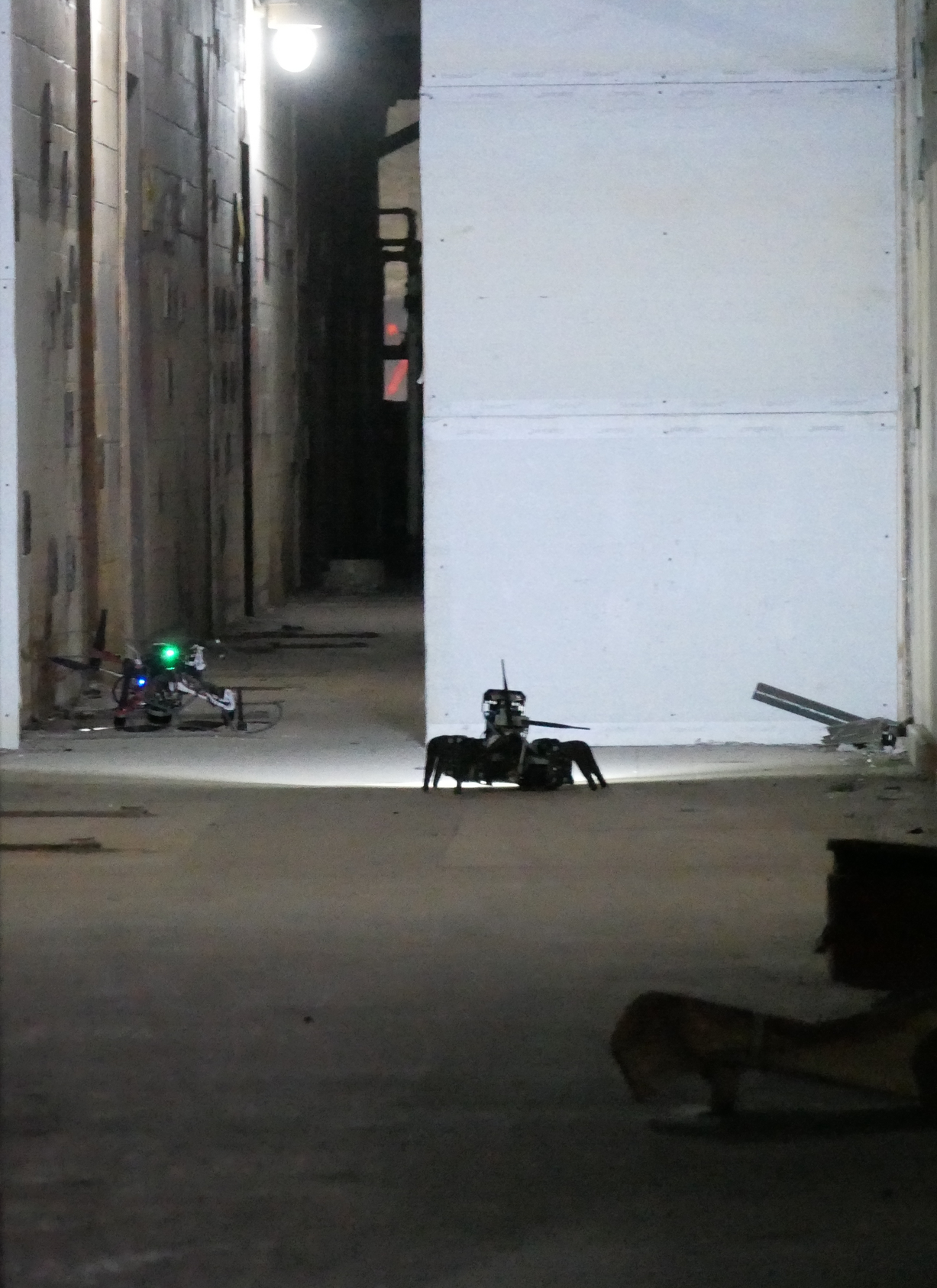}
    \caption{Team of ground robots exploring entry hallway (left) and crashed UAV being observed by hexapod (right) during DARPA SubT at Satsop nuclear powerplant.}\label{pic:team}
\end{center}
\end{figure}

Recent advances of artificial intelligence, machine perception, environmental mapping, localization in combination with the availability of computational power opened new possibilities to move robotics closer to the holy grail of achieving autonomous operation in hazardous environments~\cite{Atkeson2018}. 
However, similar to other fields of science, robotics is facing a crisis of research reproducibility. 
The first cause is the cost of experiments, which quickly rises with the amount of logistic and technical issues which need to be solved for any field experiments.
This also has to include new or destroyed equipment which happens rather often such as seen in \ref{pic:team}.
The high cost of field experiments causes the evaluations to be performed on datasets which were often gathered for a different purpose, thus making them either not sophisticated enough or not able to capture uncertainty and unpredictability of real scenarios.
Another reason is the complexity of the systems where most robots are comprised of a large number of smaller submodules from which are usually tested in separate.
Thus, method interoperability, compatibility and their impact on the efficiency of the whole system often remain neglected.
The last reason is that failures of experiments are often blamed on technical issues, and the reliability and robustness of the methods are not reported.
Instead, scientific papers focus on issues of accuracy or computational complexity of the individual methods rather than the reliability of the integrated systems~\cite{reproducible,benchmarks}.

With all the mentioned problems, the performance of robotic systems cannot be optimal in real-world situations which are impacted by the robustness of the deployed systems.
Robotic contests offer a potential solution to the problem. The results of many of these contests, such as MIROSOT~\cite{mirosot}, Eurobot~\cite{eurobot}, RoboTour~\cite{robotour}, RockIn~\cite{rockin} or MBZIRC~\cite{mbzirc}, show that the success depends more on reliability and interoperability than other aspects.

One of the active fields of research is autonomous exploration, where a team of robots is supposed to build a complete and accurate spatial~\cite{exploration} or spatio-temporal~\cite{stexploration} map of its environment.
Autonomous exploration is challenging not only because it encompasses self-localization, map building, path planning and motion control, but mainly because the robot has to operate in previously unknown areas.
Even though most of the problems mentioned above are sufficiently tackled in separate, the number of systems that can achieve a reliable exploration of truly unknown environments is low~\cite{stexploration}.

The need to quickly explore, find and locate objects in real-world environments is motivated by search-and-rescue (SAR) operations, where the time required to find survivors or environmental hazards is critical. 
The use of robotic technologies, which offer the ability to create detailed maps with positions of potential victims and hazardous areas can not only speed up the SAR operations but also make them safer for the rescue teams.  
Moreover, robots can be used to search in locations inaccessible by humans, such as gas-filled tunnels, unstable structures, large crevices, or contaminated areas.
The ability to create accurate maps can also mitigate the danger of mining disasters by preventing accidental breaches into abandoned tunnels that might be flooded or otherwise
contaminated~\cite{huber2006automatic}.

Search-and-rescue missions often have the robots controlled manually via a radio link since their ability to operate semi-autonomously is severely limited.
Even modern rescue robots offer only limited features such as intelligent camera switching or automated stop at negative obsatcles. 
The lack of autonomy means that a single operator can control only one robot at a time and teleoperating the robot imposes a significant cognitive load. 
Moreover, the primary sensory modality for teleoperation is video, which requires a high bandwidth, which, in turn, limits the number of robots deployable in the same mission.
Furthermore, maintaining a long-range, high-quality radio link between a control station and a robot that moves indoors or underground is challenging at least.
A solution to the problem is to endow the robots with the ability to operate in a semi-autonomous manner, process the bulk of the sensory data onboard and use the radio link to report only the data that are necessary for efficient and safe mission control.
The problem of underground exploration is addressed by the DARPA SubT Challenge, which offers an ideal situation to test the robustness and reliability of robotic systems in search and rescue scenarios. 

The DARPA SubT Challenge is organized not only to fund but also motivate the development of robotics systems that are capable of supporting search and rescue operations without any supporting infrastructure in the underground of post-disaster environments. 
The contest emphasizes the reliability and efficiency of complete integrated systems rather than the efficiency of the individual modules and components.
In particular, the performance of robotic teams is evaluated by their ability to quickly and accurately locate relevant objects in underground sites with a variable degree of complexity.

In this field report, we describe the multi-robot system developed for the DARPA SubT challenge by the team CTU-CRAS-Norlab of the Czech Technical University in Prague and Universit{\'e} Laval. 
Later on, we describe more details about the hardware of the robots, communication solution, localization systems, mapping, navigation, as well as our approach to multi-robot coordination.
\section{Related work}\label{sec:related}

Although there are multiple stages of the disaster management cycle~\cite{Murphy-MIT-2014}, our primary focus is on the first one: ``rescue activities during or in the immediate aftermath of a disaster to save lives''.
Robotic systems were used in several first stage scenarios such as searching for survivors after the September 11, 2001 attacks, or Hurricane Katrina in 2005. 
However, the majority of such deployments was teleoperated~\cite{Kruijff-AR-2015}.
The reason is that the robustness and explainability of the high-level autonomy methods are still far from being acceptable by rescuers.

However, adverse environmental factors in underground environments negatively impact not only the cameras but also other sensors used for teleoperation. 
Additional factors like radio bandwidth limitations, data outages and delays, and the need for fast mission execution imposes significant stress onto the robot operator, which can result in performance deterioration with severe consequences.
These factors virtually prohibit the possibility to operate larger teams of robots simultaneously.
This limitation motivated the research in autonomous and semi-autonomous exploration of environments with conditions adverse to radio communication and led to the organization of the SubT Challenge.

The work of~\cite{murphy2009mobile} summarized the impact of mine accidents and stressed the potentials of deploying robots in underground search and rescue operations.
Later works, e.g., ~\cite{neumann2014towards} investigated the use of 3D laser scanners for precise mapping, and others demonstrated the feasibility of systems that can venture hundreds of meters into abandoned mines autonomously.
Automated systems capable of creating 3D maps of underground environments were presented several years ago, e.g., in~\cite{Thrun03} and~\cite{huber2006automatic}. 
While both of these systems could create detailed and extensive 3D maps, the robots were controlled manually, and neither autonomous exploration nor object detection was considered.  
Shortly after these proof-of-concept systems, more authors addressed the problem of underground mapping.
An introduction to the problem of underground mapping along with a comparison of underground mapping systems was provided in~\cite{morris2006recent}.  
The authors of~\cite{morris2006recent} conclude that the sensor fidelity, data processing methods and mapping algorithms were not ready to constitute a reliable system capable of autonomous exploration and 3D mapping.

However, the need for reliability of underground operation requires robust solutions capable of withstanding explosions and fire~\cite{wang2014development} as well as operation in flooded mines~\cite{Maity2013}.
These often rely on pre-installed infrastructures like ceiling lights, radio beacons or magnetic wires in combination with reactive navigation behaviours~\cite{Rusu2011}.

However, development of low-weight, high-performance computers and solid-state lidars as well as progress in localization, mapping and learning algorithms allowed to integrate various degrees of autonomy into robotic systems.
Many of the works focused on mobility over adverse terrain, e.g., the authors of~\cite{zhao2008research} proposed to use a marsupial setup consisting of a ``mother'' and ``baby'' robot, which is released to map hard-to-access areas.  
Domain transfer techniques such as GANs simplified transfer from the simulator to real platforms~\cite{Pecka-RAL-2018} allowing for autonomous control of flippers on skid-steer robots.
Moreover, significant progress in reinforcement learning and trajectory optimization techniques allowed for robust control of legged platforms like ANYmal~\cite{Hwangbo-RAL-2017}.%
Reduced dimensions of sensors and computational hardware allowed to deploy real-time SLAM in aerial platforms, making them capable of operating in GPS-denied environment for Search and Rescue scenarios~\cite{Faessler-JFR-2016,petrlik2020ral}.

The availability of off-the-shelf GPUs allowed for real-time processing of high-resolution images by deep convolution networks~\cite{yolo}, which are known to achieve human-comparable accuracy in many tasks such as object recognition which is vital for visual based search and rescue.

\rnb{
DARPA's SubT directly ties back to most requirements in search and rescue scenarios and motivates multiple teams to perform their best.
For the Tunnel and Urban circuit of the challenge, each team brought its approach to the problem.
Team Explorer utilizes custom UAVs and UGVs with great terrain handling capabilities due to their size, which allows them to efficiently drive through uneven terrain where smaller robots would struggle.
The bulk of the robots allows them to carry and drop many communication nodes throughout the course, ensuring robust communication links during the deployment~\cite{tatum2020communications}.
The teams experience with underground exploration robust platforms allowed them to win the first Tunnel circuit.
}

\rnb{
Team CoSTAR focuses on their NeBula (Networked Belief-aware Perceptual Autonomy) system, which takes into account the uncertainty of all possible parts and is designed to perform long tasks outside of
the range of communications.
In \cite{bouman2020autonomous} they focus on new Boston Dynamics platforms, which with the help of two Husky platforms, won the entire Urban circuit.
Legged platforms were deployed earlier in the Tunnel circuit by Pluto~\cite{miller2020mine} and Cerberus~\cite{bjelonic2020rolling} where the legged platforms did not perform well due to slippery
surfaces even when equipped with wheels to serve as walking/rolling platform.
Other rather unconventional methods of locomotion were explored by team NCTUs~\cite{huang2019duckiefloat} which deployed a blimp.
}
\rnb{
\subsection{Contributions}
This paper provides an overview of a system utilised by CTU-CRAS-NORLAB during the SubT Challange. This system builds upon the experiences of the aforementioned works and on the lessons learned
during projects aimed at robotic search and rescue missions~\cite{tradr}. Specifically we provide our contributions as:
\begin{itemize}
 \item Robust multirobot system with redundant communications  (Section \ref{sec:software} \& \ref{sec:robots})
\item Visual, radio and olfactory based object localisation (Section \ref{sec:detection}) 
\item ICP based localisation and mapping (Section \ref{sec:localisation})
\item Adaptive traversability (Section \ref{sec:navigation})
\item Elevation mapping and entropy based exploration   (Section \ref{sec:exploration})
  \end{itemize}
}

\section{Contest specification}

DARPA's Subterranean Challenge (SubT) is a contest organized by the Defense Advanced Research Projects Agency (DARPA).
In the SubT Challenge, a team of mobile robots has to actively search an environment to locate and report the positions and types of specific artifacts in a previously unknown, subterranean environment with
limited human interaction. 
The performance of the robotic team is evaluated by the number of objects detected, and the accuracy of their position estimation ~\cite{DARPArules}.
\rna{
All participants have to adhere to the same rules, which are evolving throughout the whole competition, based on previous observations or sites where each circuit takes place.
Moreover, running in parallel is a virtual component of the same challenge, which takes place solely in a simulation.
}

\rna{
\subsection{Competition rules}
With limited time (usually 30 minutes) and personnel (maximum 10 people), each team has to prepare and then deploy a robotic system to explore a previously unknown environment inaccessible to the personnel.
The robotic fleet has a mission to find, locate and then report back to base the position of the previously given objects called artifacts within a limited time (usually 1 hour).
While it is not required by the rules for a system to be fully autonomous, only one person is allowed to control any aspect of the mission and view the data reported back by the fleet.
\subsubsection{Score}
All artifacts that have to be found change for each circuit and are specified by DARPA beforehand.
For a team to obtain a point, it is required to
\begin{itemize}
\item find and recognize the correct type of artifacts,
\item localize the artifact with a maximum of 5m of euclidian distance from the ground truth in the global coordinates,
\item and report the correct type and position back to home base.
\end{itemize}
  There are 20 artifacts on the course. Each team can send 40 reports during one run, making it not possible to brute force the problem.
  Additionally, each artifact has to be reported only once, otherwise a point is subtracted from the team.
  If any teams would have the same number of points, other factors such as how fast the artifacts were reported or how far teams traveled are then used as a tie breaker.
  More information can be found on the DARPA SubT website~\cite{DARPArules}.
}
\begin{table}[!ht]
\begin{center}
   \caption{Summary of DARPA organised events for SubT with their environments and artifacts for each round. Artifacts from SubT Integration Exercise (STIX) are common for all Circuits~\cite{DARPArules}.} %
   \renewcommand{\arraystretch}{1.2}
   \begin{tabular}{l|l|l|l}
     Event & Date&Enviroment & Artifacts \\
     \hline
     STIX Exercise& 04/2019&Silver Mine & Backpack, Survivor, Cellphone \\
     Tunnel Circuit & 08/2019 & Coal Mine&AKU Drill, Extinguisher\\
     Urban Circuit & 02/2020 & Nuclear Plant&CO2 Elevation, Duct Vents\\
	   Cave Circuit & Canceled &Natural Cave&Rope, Helmet\\
	   Final Event & 09/2021 & Louisville Cavern& All above, Translucent SubT Logo\\
\end{tabular}
\end{center}
\end{table}

\rna{
\subsection{Contest Challenges} 
The whole Challenge is split into four consecutive rounds, called ``circuits'' where each round takes place in a specific domain (such as ``Tunnels'', ``Urban'' and ``Caves'') and combines them all in the final round. 
Each of these environments is built to provide various challenges and serves as an incubator for new technologies and approaches in several robotics problems such as limited communications, traversability, multiagent integration, sensing, etc.
}

\section{Robots and Sensory Equipment}
\label{sec:robots}

Robots utilized by our team include wheeled and tracked unmanned ground vehicles (UGV), six-legged spider-like robots, and unmanned aerial vehicles (UAV).
This usage of several types of robots is crucial since each platform has different capabilities.
Larger robots with wheels are able to traverse vast distances since they are the fastest platform even compared to tracked robots which are capable of traversing difficult terrain where the wheeled robots would get stuck.
The legged robots and quadrotors do not carry the same equipment as larger robots but can access narrow tunnels, elevated vantage points, and places that would otherwise be inaccessible by larger
robots.
\rna{
	The specific equipment carried by each platform type varied thoughout the development but the general types of sensory and computation equipment stayed mostly the same, see Table~\ref{tbl:payload}.
\begin{table}[!ht]
\begin{center}
	\caption{\rna{Hardware information and payload of each robot type.}}\label{tbl:payload}
   \renewcommand{\arraystretch}{1.2}
   \begin{tabular}{c|c|c|c|ccccccc}
     Robot & Husky A200 & Absolem & Hexapod& UAV\\
     \hline
     Movement & Wheeled & Tracks& Legged(6)& Quadrotor\\
     Speed [m/s] &1&0.4 & 0.2 & 1\\ 
     Traversability & Low& Medium & Medium & High \\ 
     Computation power& High & High & Low & Medium \\
     Max. payload [kg]& 50 & 10 & 1 & 0.4\\ 
     Runtime& 1.5h & 1.5h&1h&10-20m (payload dependent)\\ 
     Communications& All & All & Mote& Wifi,Mote \\ 
     STIX&0&2&0&2 \\
     Tunnel&1&2&2&2\\
     Urban&1&3&1&2\\
\end{tabular}
\end{center}
\end{table}
}

\subsection{Wheeled robot - Husky A200}

The Husky A200 from Clearpath Robotics is a differentially driven wheeled platform.
The platform is built to be rugged and capable of traversing mud, gravel, light rocks, and steep declines/inclines.
The onboard sensors changed between circuits, consisting of lidar with a $360^\circ$ field of view and a range up to 200~m, which is the primary sensor for localization and mapping. 
In the``Tunnel'' circuit we used a Robosense lidar with 32 lines, and in the ``Urban'' circuit we used a Robosense 3D lidar with 16 lines. 
To detect objects of interest, the Husky is equipped with 5 Bluefox RGB cameras positioned to achieve a 360$^\circ$ field of view and one Bluefox camera to look upwards.
All the cameras are running at a reduced 15~FPS with extended exposure to allow for low light operations.
The camera feed is then passed to the onboard NVidia Jetson TX2, which is used for object detection.
Calculations for localization, mapping, and control were performed on a separate control PC, Intel NUC5-i5 on the ``Tunnel'' and Dell OptiPlex 3070 on the ``Urban''.

Since the payload capacity far exceeds the weight of the sensory and computational equipment, the robot carries some extra devices.
In the ``Tunnel'' circuit, it was equipped with two eight-slot containers for communication relays.
In the `` Urban'' circuit, the design of communication relay containers was changed, so the Husky carried only four relay modules.
These relay units are deployed during the mission to extend the range of low-bandwidth communications from the command station~\cite{cizek_mesas}.
Other possible attachments are docks for legged robots and UAVs, which could be deployed on the most suitable locations for those specific platforms~\cite{uav_ugv}.
\begin{figure}[!htb]
\begin{center}
    \includegraphics[width=1.0\columnwidth]{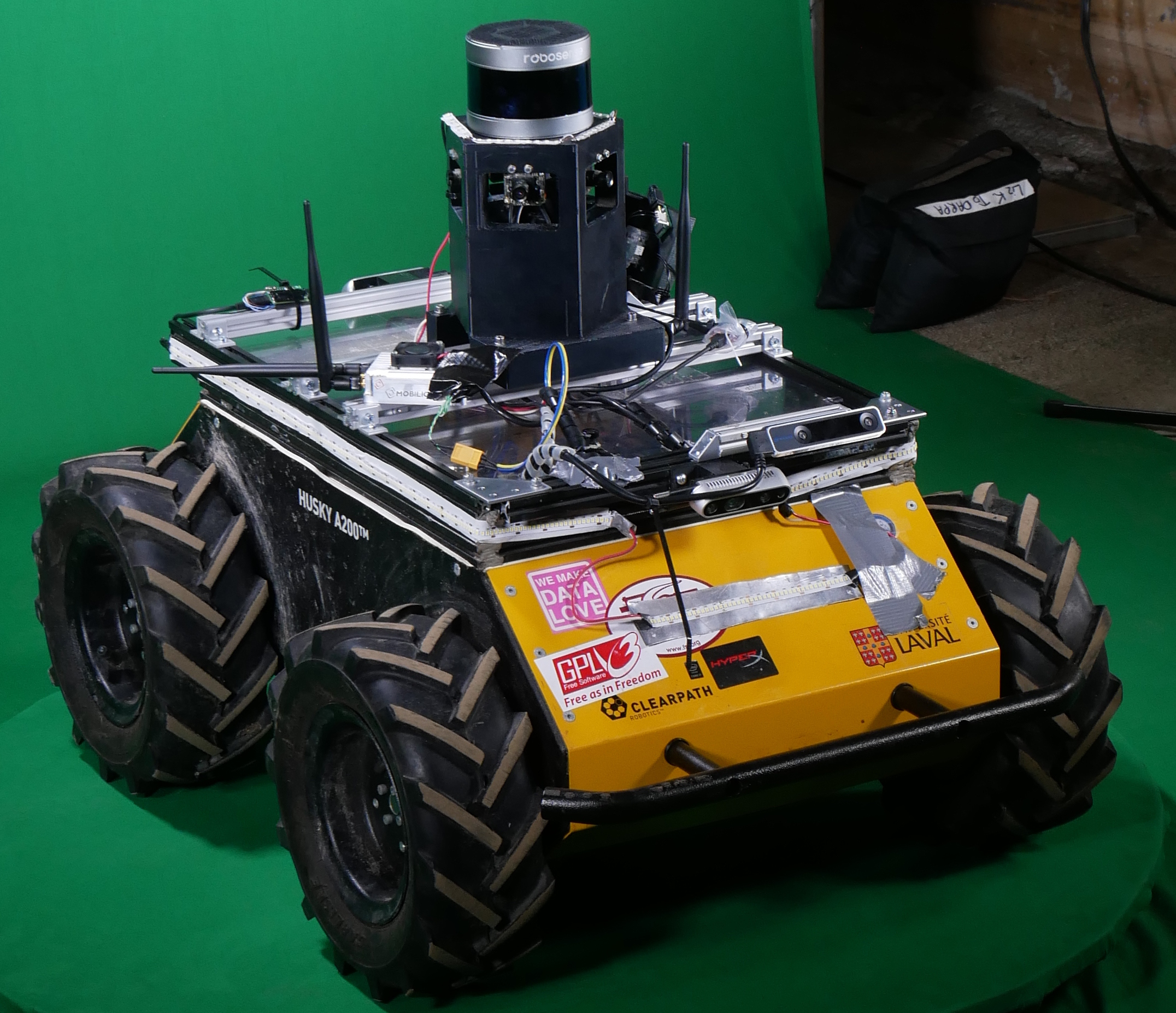}
    \caption{The Husky platform with a Robosense lidar on top of a sensory tower which houses five cameras.\label{fig: Husky}} %
  
\end{center}
\end{figure}

\subsection{Tracked robots - Absolem}
The three tracked Absolem platforms made by BlueBotics SA are designed to traverse terrain typical of disaster sites.
All Absolem robots drive on two main tracks, each having two additional sub-tracks called flippers, which are controlled independently during the drive of the robot.
The flippers are set by a custom controller, which takes the flipper position and data from mounted sensors to assess the best setting for those flippers to help the robot traverse large obstacles such as
stairs, curbs, and even gaps larger than the robot itself.
The primary sensor for localization and mapping is a SICK LMS-151 line lidar mounted on the robot's front, pivoting around its forward axis.
The lidar is attached to a motor that rotates it. This makes the scanning plane sweep the space, making it possible to acquire a 3D point cloud of its surroundings.
Processing power is provided by an internal PC with Intel Core i7, which runs localization, mapping, and navigation.
PointGrey Ladybug 3 omnicamera (now FLIR Ladybug 3) is used to feed 4K pictures at 6~FPS to a Jetson TX2 board for object detection and camera feed.
As the computational demands of our algorithms increase, for the ``Urban'' circuit, we have also added an offloading computer (Intel NUC8-i7) which can process the mapping and other CPU-intensive operations.
The Absolem robots are supposed to thoroughly search areas not accessible by the Husky robot, e.g., those accessible only by stairs.
This is possible due to the tracked design and size, which allows for enough sensors to be mounted to enable the same algorithms to be run for the Husky and the Absolem.
This software will be described further on in Section~\ref{sec:software}. 

\begin{figure}[!htb]
\begin{center}
    \includegraphics[width=1.0\columnwidth]{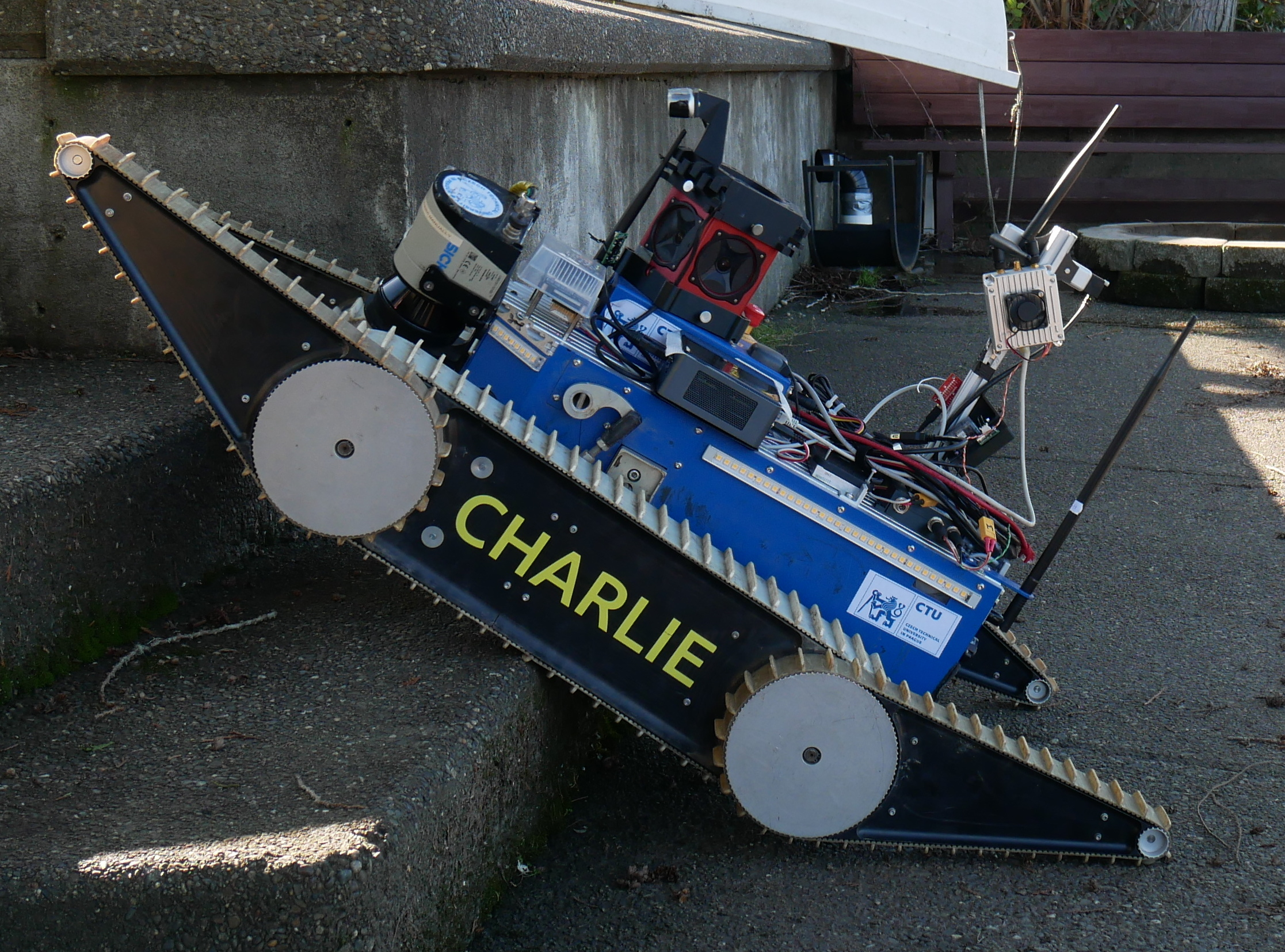}
    \caption{Absolem climbing concrete stairs using extended flippers.\label{fig:tradr_solo}}

\end{center}
\end{figure}
\subsection{Crawling robots - hexapods}
The crawling spider-like robots are derived from the six-legged PhantomX Mark II platform.
Navigation, mapping, and localization are done by a custom rig comprised of two  Intel Realsense cameras supported by LEDs to compensate for low light conditions.
For localization, T265 is used since it can run onboard SLAM, providing an estimate of the robot's actual position.
The onboard SLAM makes it possible to lower the computational load of a robot's computer~\cite{bayer19ecmr}.
A D435 RGBD camera does mapping, exploration, and object detection.
Computation is done on a NVidia Jetson TX2, which is the only computer on those platforms, mainly because it can perform object detection.
The main strong point of the hexapods is an ability to crawl over hard terrain or through vents or ducts that would be otherwise inaccessible to other ground robots~\cite{faigl2016localization}.
Since their speed is rather slow, they are not much use except for some small extensions as communication nodes.
This can be are largely overcome by potentially supporting them with Husky robot to carry them to the places of interest during the competition.
\begin{figure}[htb]
\begin{center}
    \includegraphics[width=1.0\columnwidth]{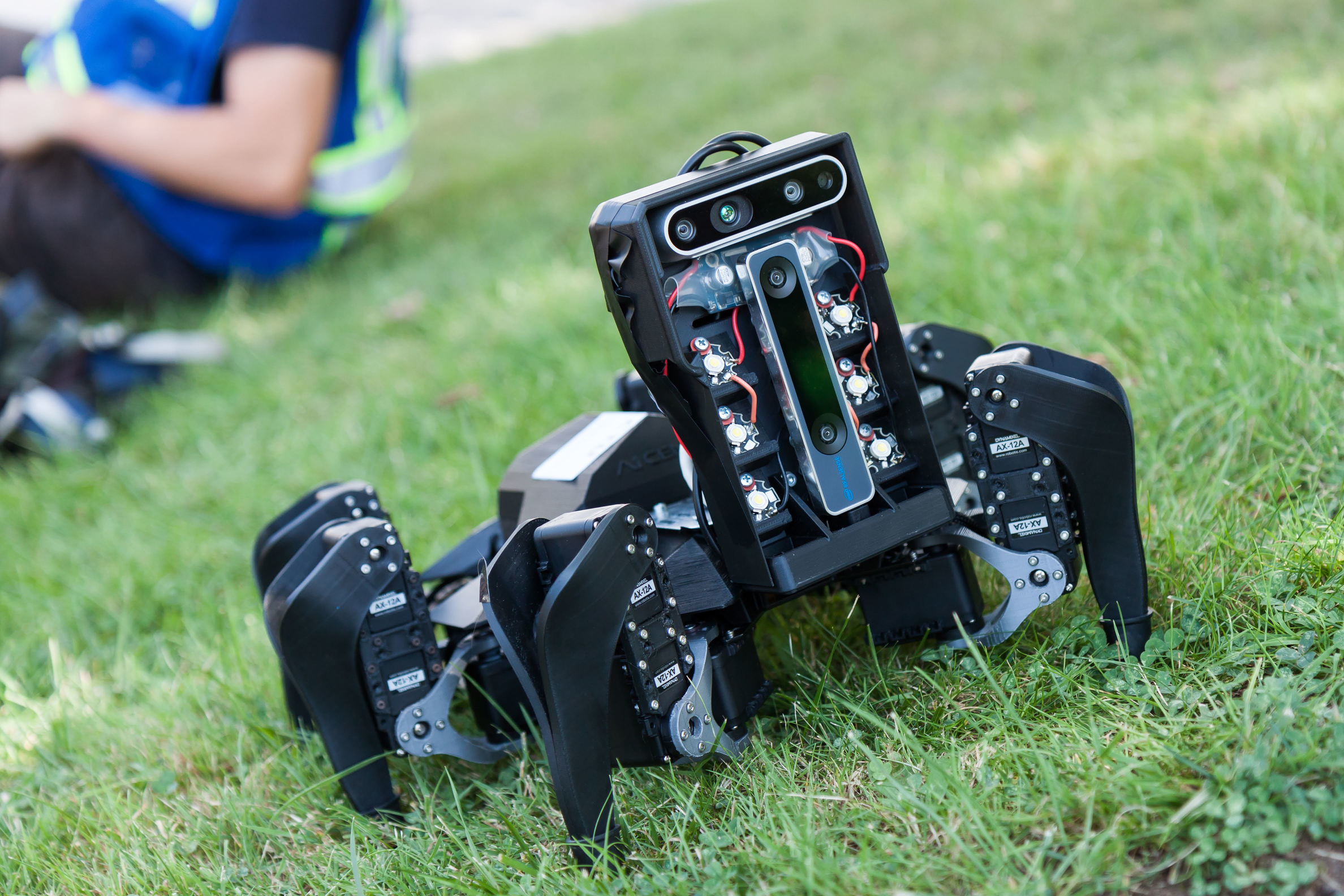}
    \caption{Hexapod in grass ``sitting'' in power save mode.\label{graph:hexapod_solo}}

\end{center}
\end{figure}

\subsection{Aerial robots - quadrotors}
The aerial robots are based on the F450 kit by DJI, controlled by an onboard Intel\textsuperscript{\textregistered} NUC and the PixHawk autopilot.
Their primary sensor to perform localization and mapping in the ``Tunnel'' circuit~\cite{petrlik2020ral} was RPLidar A3 2D lidar, which provides 16000 distance measurements per second in a 360$^\circ$ circle in one plane.
The sensor range is \SI{25}{\metre}, and its (adjustable) rotation rate is set to \SI{10}{\hertz}, which provides sufficient overview to safely move in narrow tunnels with speeds of up to \SI{1}{\metre/\second}.
Laser-based localization was done by the Hector SLAM~\cite{hector} package, which is supposed to work well in conditions that can be anticipated in the subterranean
environment~\cite{santos2013evaluation}. 
In the Urban circuit, we have upgraded to a 3D lidar Ouster OS1-16, which provides $360^\circ$ scans in 16 beams. This greatly improved the mapping and localization capabilities of the UAVs.
For object detection, each UAV carries at least two Bluefox RGB cameras with an LED illumination strip.
The detection itself is performed by a custom trained YOLOv2 CNN on the CPU of the onboard i7 NUC PC, which also performs localization and mapping.
With this setup, the processing of one image on a single thread takes approximately \SI{1.7}{\second}, which is just sufficient in order not to miss an object when flying at standard velocities.  
Both UAVs are set up to automatically run the exploration process after launch to be deployed easily by any member of the team.  

\begin{figure}[htb]
\begin{center}
    \includegraphics[width=1.0\columnwidth]{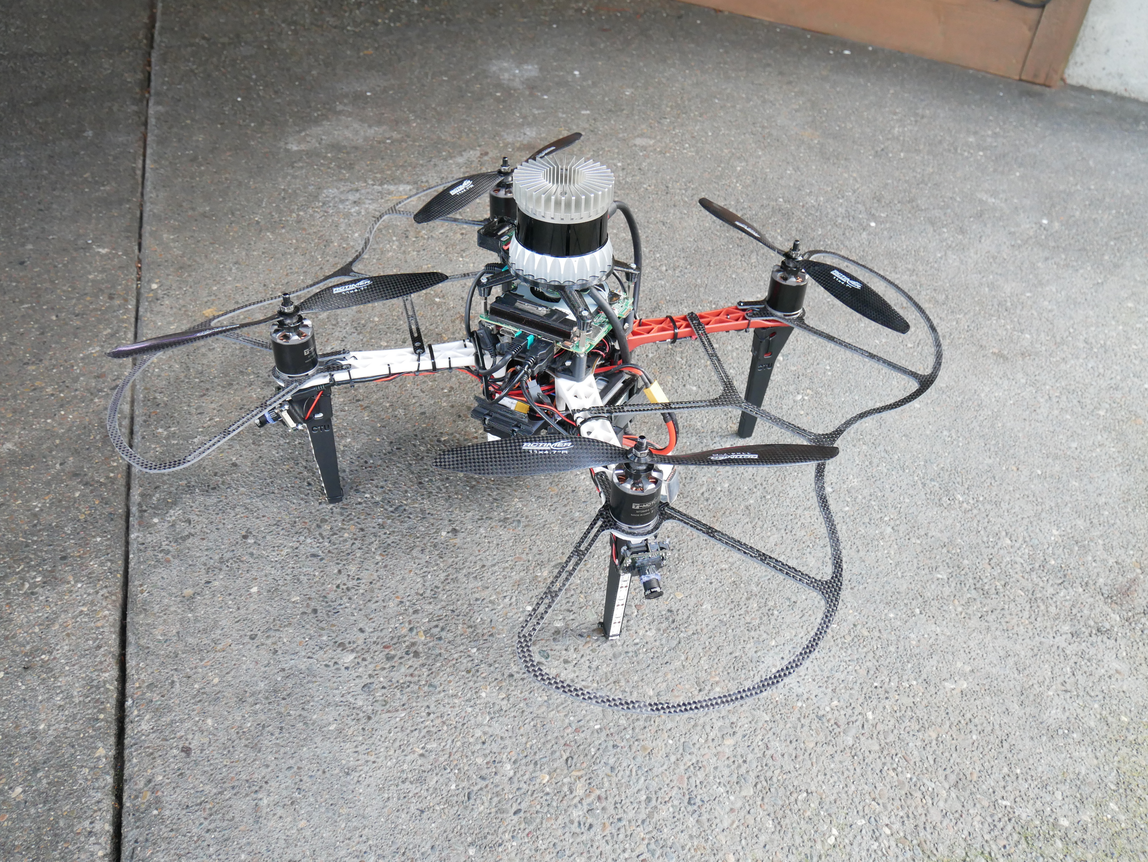}
    \caption{One drone platform with propeler guards and Ouster sensor on top.\label{graph:drone_solo}}

\end{center}
\end{figure}

\subsection{Communication}
\label{ssec:communication}
Of the previously mentioned problems, one of the most limiting is the communications constraints.
During search and rescue situations, interweaved hallways, underground tunnels, shielded rooms, and thick steel-reinforced concrete walls make it hard or even impossible to maintain any stable radio
link.
This prevents direct teleoperation by the operator, which otherwise could substitute for complex decision-making algorithms that are required for any autonomy.
In the ideal case, the operator would have no constrained link to the robots, which would provide him with all the information about their surroundings.
Alternatively, robots would be fully autonomous, requiring no communications eliminating the need for the operator.
Autonomy is hard to do reliably, and even fully autonomous robots would have to share some of the information back to the operator, so our team implements several different communication systems to
deliver at least some data to the human supervisor of the system. 
These solutions are all redundant to a certain degree.
Losing one of the connections usually does not mean any connection at all but rather an unstable link.
This can manifest as lower quality and framerate of the video, fewer status reports, or just position reporting without any other data. 
All of them have different levels of reliability, range, bandwidth, and usage when it comes to their specifications.
All robots were also equipped with the capability to be connected by wire using ethernet for setup.

\begin{table}[!ht]
\begin{center}
  \caption{\rnc{Communication methods comparison.}}
\label{fig:coms-table}
\begin{tabular}{c|ccc}
  
     Type & WiFi & Mobilicom & MOTE \\ \hline
     Range& Short&Medium &Large\\ \hline
     Fequency & 5GHz &2.3GHz& 900MHz\\\hline
     Bandwidth & 50 Mbit/s&1 Mbit/s & 432 bits/s\\\hline
  Size[cm] & Varies& 8x8x3 &\shortstack{2x3x0.5 on robot + antenna \\ 4x4x10 dropped} \\
\end{tabular}
\end{center}
\end{table}

\subsubsection{Short-range link: WiFi}
During testing and usual usage, the robots are connected to standard Wi-Fi which has a bandwidth that allows for the transfer of high data-usage information such as direct sensor measurements or real-time
video feed.
This link is used during the initial setup of the deployment, where all the robots are in close range of each other since it is necessary to verify readiness for the mission and do the initial setup.
Later on, robots use the same Wi-Fi link to send data back to the base station, but the signal deteriorates quickly, and the subsequent mid-range communication needs to take over.
Additionally, the robots can use the Wi-Fi to share other gathered information such as 3D maps or past positions with others and can use this link in case of failure of the others.
When a UAV was sent to the mission without the Mobilicom unit (described below), it used Wi-Fi to connect to the base station when it returned to report its findings.

\subsubsection{Mid-range link: Mobilicom}
During the actual mission or large-scale testing, special units called Mobilicom are used.
These modules use high-power transmission with dual-channel capability and automatic meshing abilities to construct a network that can send data over much larger distances than Wi-Fi.
During runs, robots are coordinated to provide retransmission nodes to maintain communication with all the robots at all times.
This makes a network that is able to dynamically switch to the best possible path between each unit to send valuable information for most of the structures we have yet tested.
Since the data throughput is limited to 1~Mbit/s, rationing this to robots proved to be a challenging task but allowed us to have a low-quality video for teleoperation, basic environment maps, and object
detection pictures to be transmitted through the network.
In the ``Tunnel'' circuit, we used the larger MCU-30 Ruggedised units and MCU-200 with higher transmission power as the base. The size of these modules allowed their use on the wheeled and tracked robots only, which limited the system to a communication range of roughly 300~m. 
In the ``Urban'' circuit, the smaller MCU-30 Lite units were used, and this allowed the deployment of the communication nodes on all robotic platforms, including UAVs and hexapods, to help extend the network range.

\subsubsection{Long-range link: Motes}\label{sec:motes}
To have more redundancy and coverage, one more communication module type is used.
These small custom-built modules allow for low bandwidth, which is sufficient to share crucial data from the robots, such as their
status, position, and possible detections.
The data are not shared only to the base station but also to the other robots to enable low-level coordination and control from the base station.
These ``Mote'' modules are carried by all our robots, even UAVs, since they are compact and small.
The larger robots can even carry several of these modules and can drop them to create a network of relays, providing communication infrastructure for the other robots. 
In the ``Urban'' circuit, due to the code and bandwidth improvement of these modules, we were also able to send commands to the robots from the base station via the Motes.
This helped in cases where the robots went out of the range of the Mobilicom nodes, and the operator wanted them to return to the signal. 

\section{Software}\label{sec:software}

While appropriate hardware is necessary for successful participation at the challenge, it is the software that brings the entire system alive.
To stress the importance of the software, DARPA SubT has a parallel ``virtual track''  where the contest occurs only in software simulations of the deployment sites.
While not all the teams participated in the ``virtual track'', most of them utilised the simulators as an integral part of the development process since tests in a simulation are much less tedious than real-world testing.
Moreover, freely-available, open-source simulators can be used to evaluate the performance of developed software modules in isolation as well as in a holistic way.
The software modules of all robots had to tackle localisation, mapping, navigation, object detection, exploration and multi-robot coordination.
\begin{figure}[!htb]
\begin{center}
    \includegraphics[width=1.0\columnwidth]{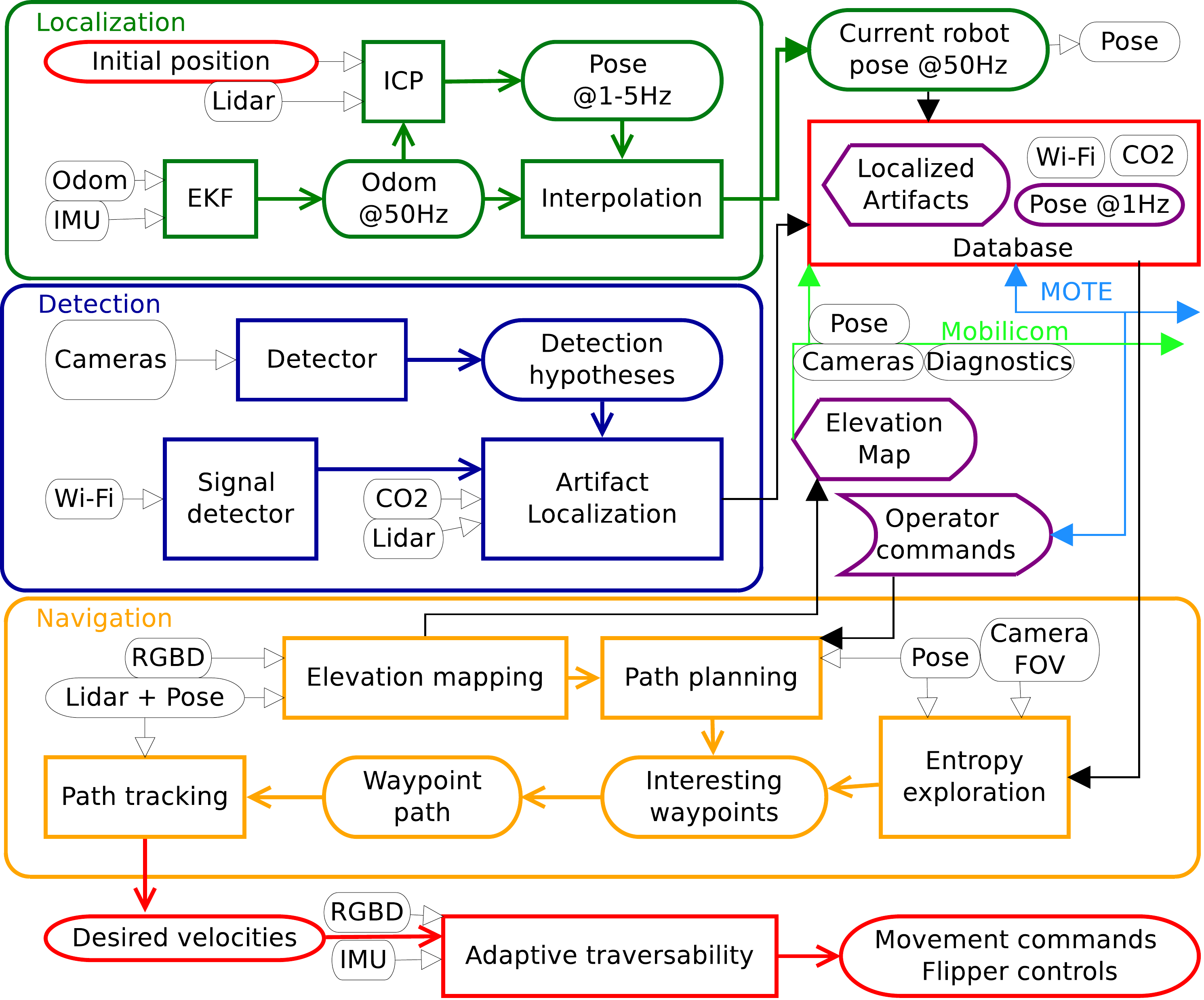}
    \caption{\rna{Block diagram of software modules on a single UGV platform with three main sub-systems: localization (green), detection (blue) and navigation (yellow).
	Purple boxes and database indicate the main I/O of a robotic platform that is transmitted to other agents via the MOTE (light blue) and Mobilicom (light green) networks.
	Internal robot states and sensory information are represented as bubbles while modules are represented as rectangles.
	Initial position is obtained from the pit crew during setup.}} 
\end{center}
\end{figure}

\subsection{Software Architecture}

All of the systems used in the competition were running the Robot Operating System (ROS)~\cite{ros} in its Melodic version on Ubuntu 18.04 or its variants (i.e., Linux4Tegra 32 on Jetson TX2 computers). %
Choosing ROS allowed us the benefit from the multitude of packages developed by the open-source community around ROS; thus, we could concentrate on the software that is specific for our platforms.
For example, the whole driver stack of the Husky robot up to EKF-based localisation, RGB camera and lidar drivers was using just the publicly available code from ROS community. 
The Absolem platforms have a free base drivers (but not open sourced), but all sensor drivers and a large part of sensor data processing are based on the code provided by the ROS community.
The community-provided packages do not, however, have only advantages.
ROS does not (yet) provide any terms of software quality measures, so it happens that as we are using a~package, we discover it is either buggy or not written in an optimised manner.
Thankfully, as all the packages are open-source, we can always fix the issues locally, and then offer the fixes to the upstream repositories.

For SLAM, we are using the open-source mapper called ``ethzasl\_icp\_mapper''\footnote{\url{https://github.com/ethz-asl/ethzasl_icp_mapping}} with customisations done by the mapping experts from our partner Universit{\'e} Laval. %
Our planning and exploration solution is not open-source, and it is built in a way that it can even work without ROS, but of course, it has a ROS interface to be able to communicate with the rest of the software modules.

One of the benefits ROS brings to us is the ease of decision where should some code be running.
As mentioned in Section~\ref{sec:robots}, most robots carry two or three computers on board.
The various types of algorithms we use and their constant development require high flexibility in terms of computational power, and with ROS, running the CPU- or GPU-intensive programs on different
computers is just a matter of a few configuration lines and some one-off provisioning. %
ROS also provides high-quality visualisation tools like RQt and RViz, which we utilise in our system both during development and for the operator console (described further in \ref{sec:interface}).

\subsection{Network Architecture}

To get a resilient and distributed communication architecture where some links may disappear during the mission, we decided not to use the standard single-master mode ROS supports.
Instead, we use a multi-master solution based on the ``nimbro\_network'' package with customisations, and also a RocksDB database.
Each robot runs a ROS master on its main computer, so on the robot's local network, all computational nodes and other supporting programs are running as in a standard single-master ROS system.

However, to connect with the base station, the multi-master solution allows us to a) explicitly state, which data can be transmitted via the limited-bandwidth connection, b) use UDP protocol where
appropriate (which helps to prevent network congestion), and c) be resilient to network failures.
As most connections between the robot and the base station are made via UDP protocol, the connection state is not affected by the actual wireless link status -- it just transmits data whenever the link is available.

The multi-master solution is only suitable for streaming-type data like camera images or 3D information; for example, the artefact detections and history of the robot's movement need to be also transmitted for the times when the link was not available (after re-establishing it, of course).
For this kind of data, all robots and the base station have a RocksDB database where they store the mission-critical data, and there is a replication protocol between each robot and the base station, via which they synchronise their databases.
This synchronistaion assures that if the robot was operating out of the reach of the wireless link, we would not miss any artefacts making him send the collected data back after the link is
reestablished. %
For the future rounds, we would like to implement peer-to-peer database synchronisation also between the robots, which would allow us to better utilise the wireless network link and make the setup more resilient.

\subsection{System resilience}

One of the key features of our software is that it is written with resilience in mind.
We know that we do not have the human resources to test all software we write.
We also know that we work with cutting-edge hardware is high stressed during the mission, and it either is not tested for our use-case (like the Realsense T265 cameras) or is just prone to fail in many different ways (screws getting loose, cables getting loose, motors getting overloaded, sensors being blinded or misreporting, and many more).

So instead of going the way industry chooses -- which means endless tests, verifications and certifications -- we try to write the software in a way that allows for failures and tries to mitigate them.
If a vital software node crashes, we restart it hoping that it will not crash again (and we investigate the crash cause after the mission).
Some nodes (like the SLAM) require some state to be restored when we restart them after a failure, so we log the state to the filesystem and let the nodes read this state and reinitialise them in the last known working state.
From the sensor point of view, we try to duplicate some of the mission-critical sensors, or, better, have several different sensing methods for each sensing modality.
For example, the flipper control algorithm needs to know the local terrain shape, so it utilises data from an RGBD camera but also takes data from the slower lidar.
When the RGBD camera fails (which it does because it is connected via USB), the algorithm still has at least some data it can work with.
One of the best achievements of these resilience techniques was that during the Urban circuit, on of our robots' main computers got completely restarted during the mission after a hard hit, and the operator was able to connect to the robot and get it up and running again in a very short time, and the robot even kept the knowledge about where it is in the map.

Some parts of the system need a more delicate approach than just restarting when it fails.
For example, if the motor that rotates the Sick lidar on Absolem platforms gets stuck, we know that we need to stop the robot immediately; otherwise, the mapping would not recover with the robot going further without any 3D data coming.
After we stop the robot, we know we can initiate the motor restart procedure, which hopefully restores the motor function.
Another example is the failsafe watchdog, which is running on the robots and watching their pitch and roll angles, and if navigation fails and gets the robot to an unsafe pose, these watchdogs perform recovery actions (either help the robot with flippers or stop it).
Knowing that all these automatic recovery procedures are useful, we also know that we cannot anticipate all error conditions and that some will probably falsely trigger these procedures.
So we give the operator the right to turn these protections off and try to get the robot to a better state manually.

\subsection{Object detection}
\label{sec:detection}
The purpose of the entire system is to detect and locate artefacts that represent potential victims or provide a cue of their location. 
Therefore, every robot of our team performs the object detection task. 
Due to the fact that colour camera is the only sensor common to all of our robots, the base of the object detection is a neural-based computer vision method called YOLOv3~\cite{yolo,yolo3}.

Since the rules explicitly prohibit training the neural network at the places where the challenge occurs, we gathered our dataset from various underground environments and trained the detector on more
than 20000 images.
\rnc{
These enviroments include university hallways and basement, STIX deployment, 'Josef' experimental mine, Tunnel deployment and Prague subway station.
All data was collected though cameras of our robots while other testing was being done on the platforms or through cameras that were temporarily dismantled from the robots.
Dataset annotation was done manually. See Table \ref{tab:dataset} for numbers of annotated objects.
}

\begin{table}
   \caption{\rnc{Dataset details with the number of annotated objects. Negative labels are the images containing only the background.}}
\begin{center}
\scriptsize{
  \begin{tabular}{ l | c  c  c  c  c   c c  c }
    Environment & Images & Negative labels & Survivors &  Cellphones & Backpacks &  Drills & Extinguishers & Vents \\ \hline
    Tunnel & 10454 & 2205 & 1010 & 1010 &  1191 & 2317 & 1869 & 0\\ 
    Urban & 10123 & 2790 & 1559 & 1109 & 1316 & 12 & 247 & 4024 \\
  \end{tabular} }
\end{center}\label{tab:dataset}
\end{table}

\begin{figure}
	\subfloat[Survivor]{%
		\includegraphics[width=0.24\textwidth]{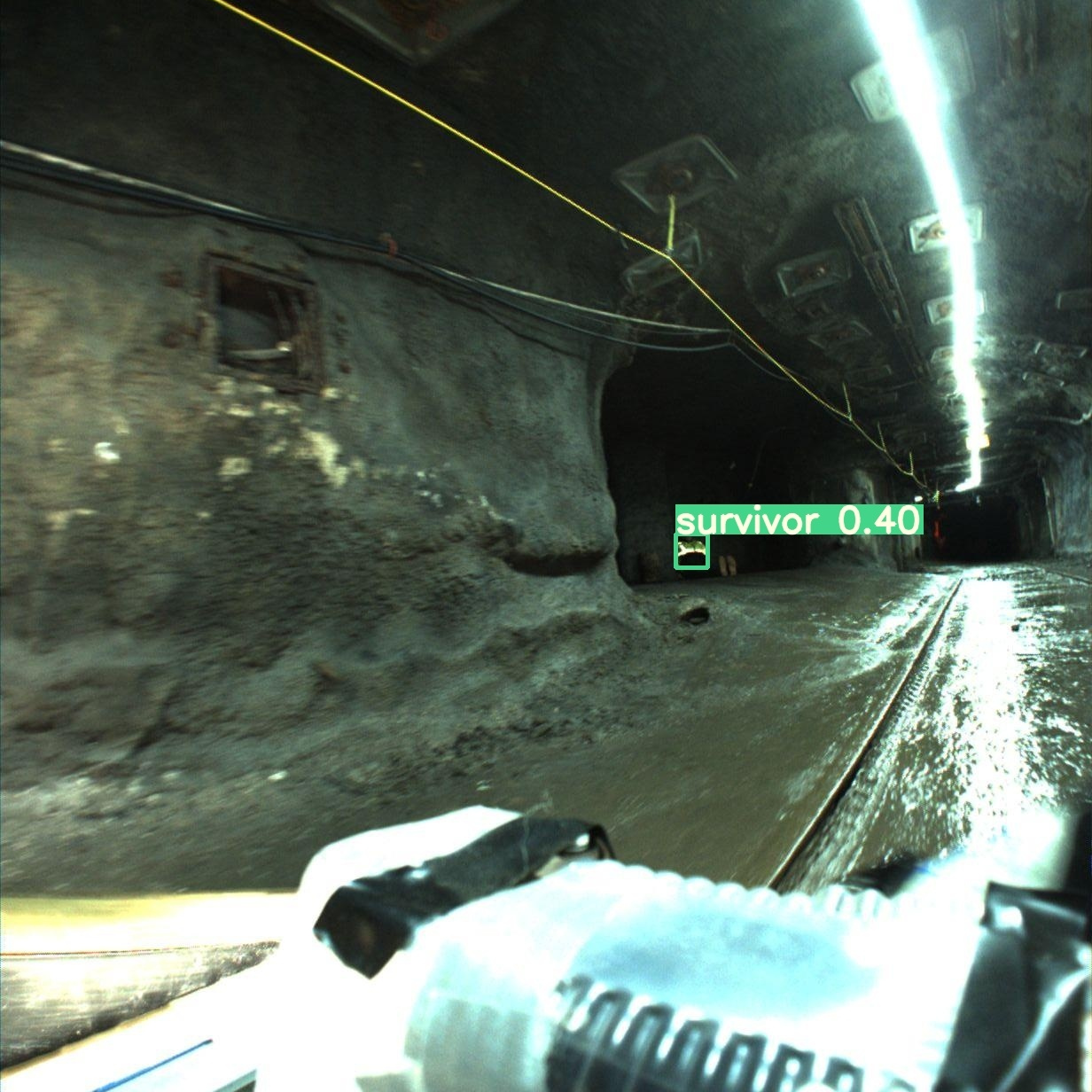}
	}
	\subfloat[Cellphone]{%
		\includegraphics[width=0.24\textwidth]{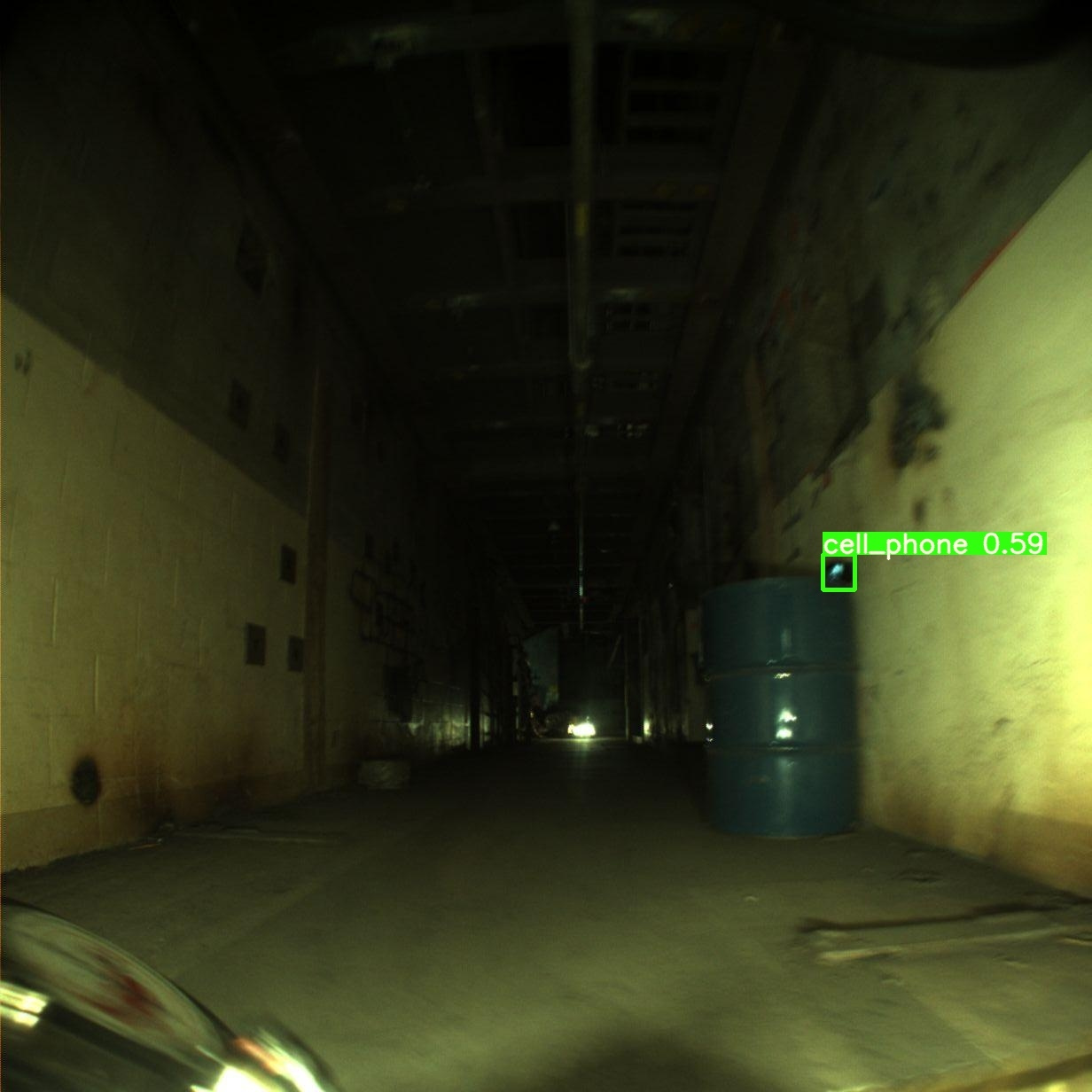}
	}
    \subfloat[Backpack]{%
    	\includegraphics[width=0.24\textwidth]{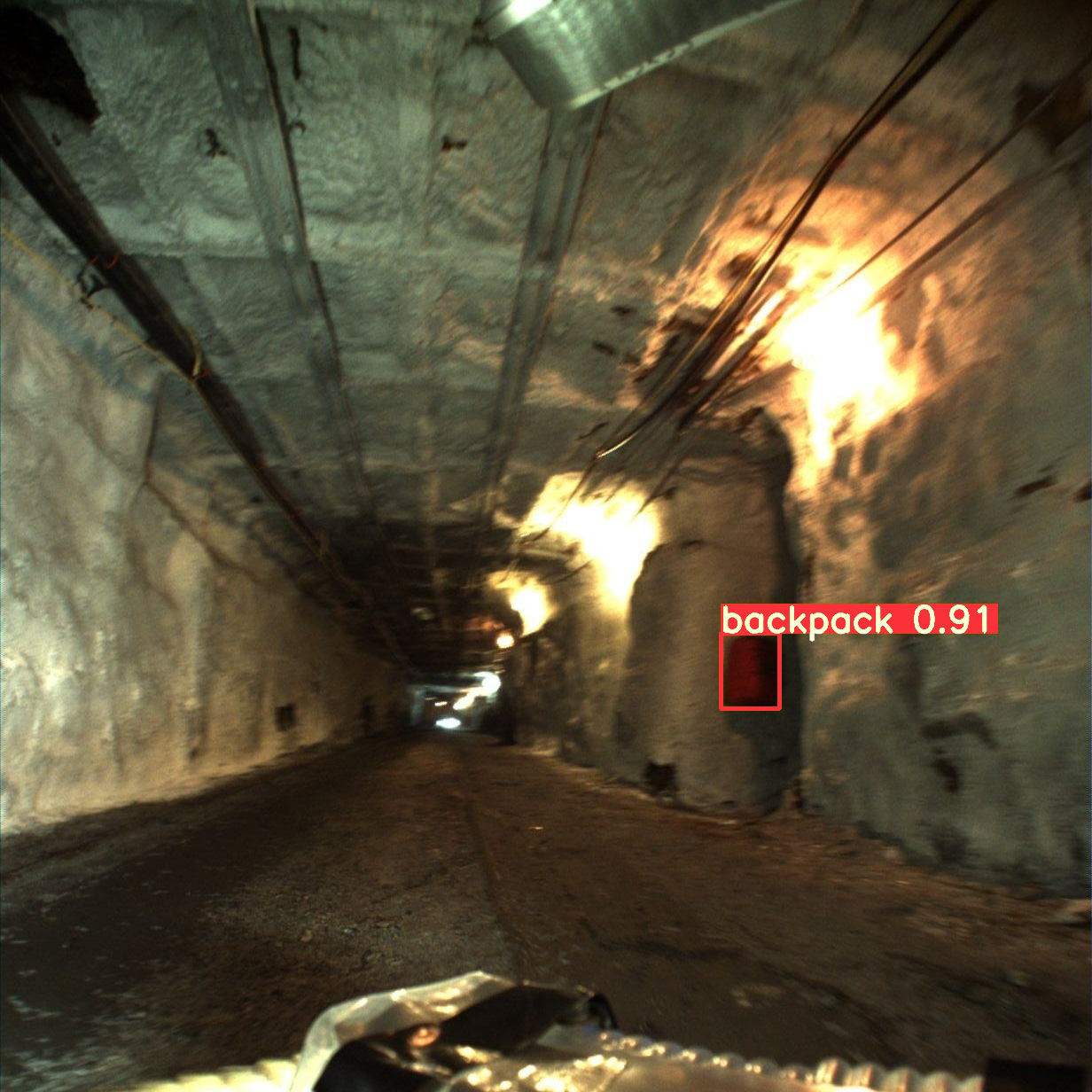}
    }
    \subfloat[Drill]{%
    	\includegraphics[width=0.24\textwidth]{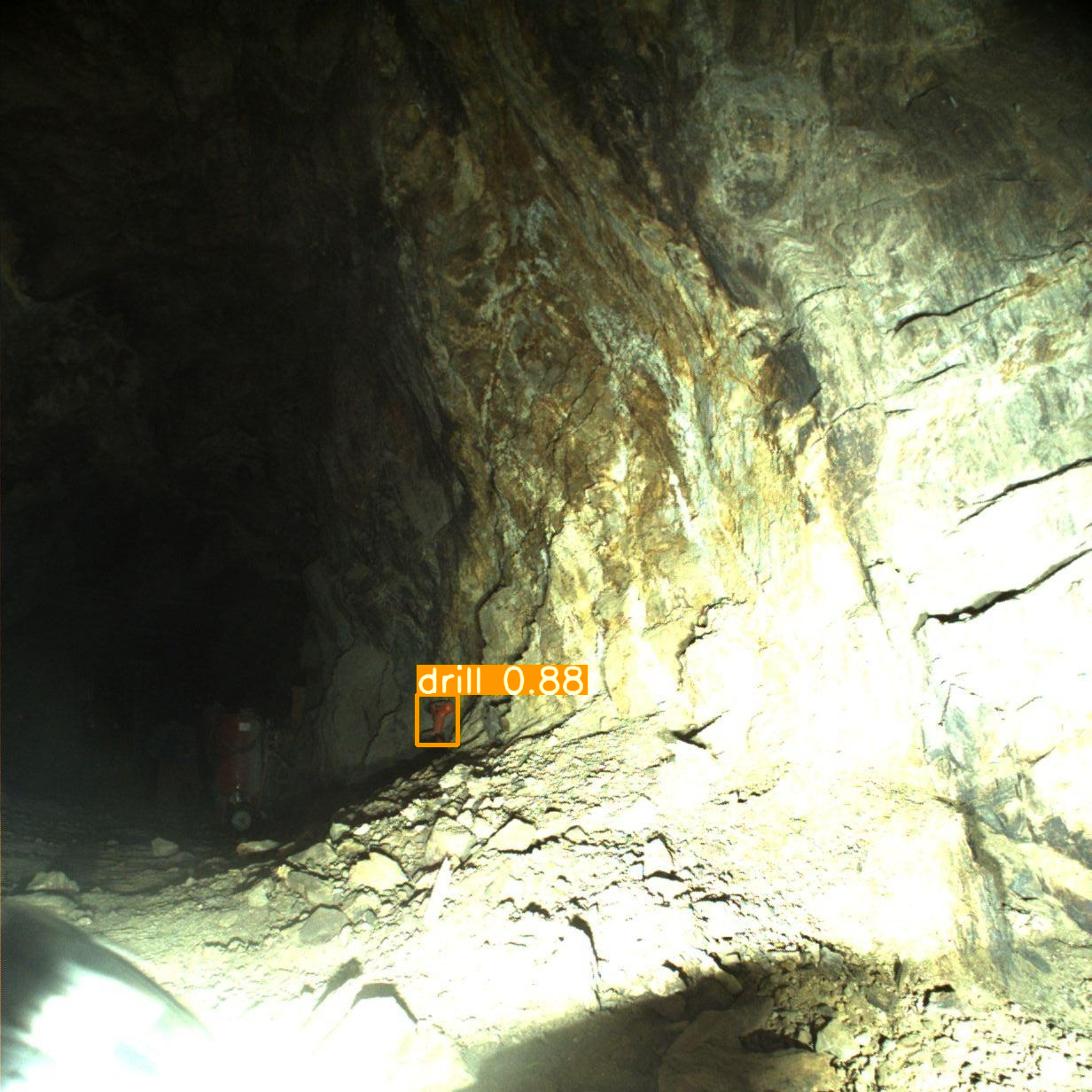}
    }\\
	\subfloat[Fire extinguisher]{%
		\includegraphics[width=0.24\textwidth]{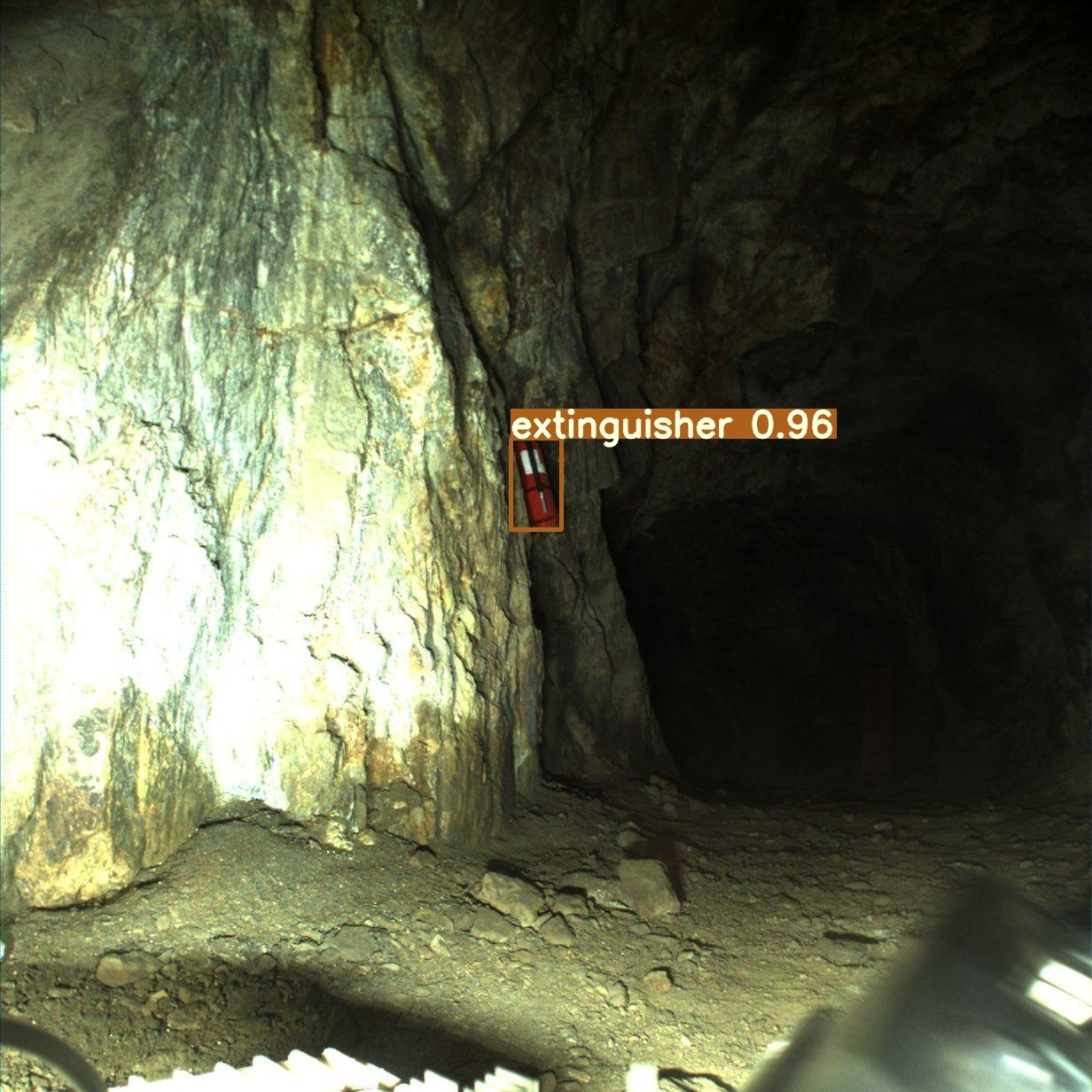}
	}
	\subfloat[Vent]{%
		\includegraphics[width=0.24\textwidth]{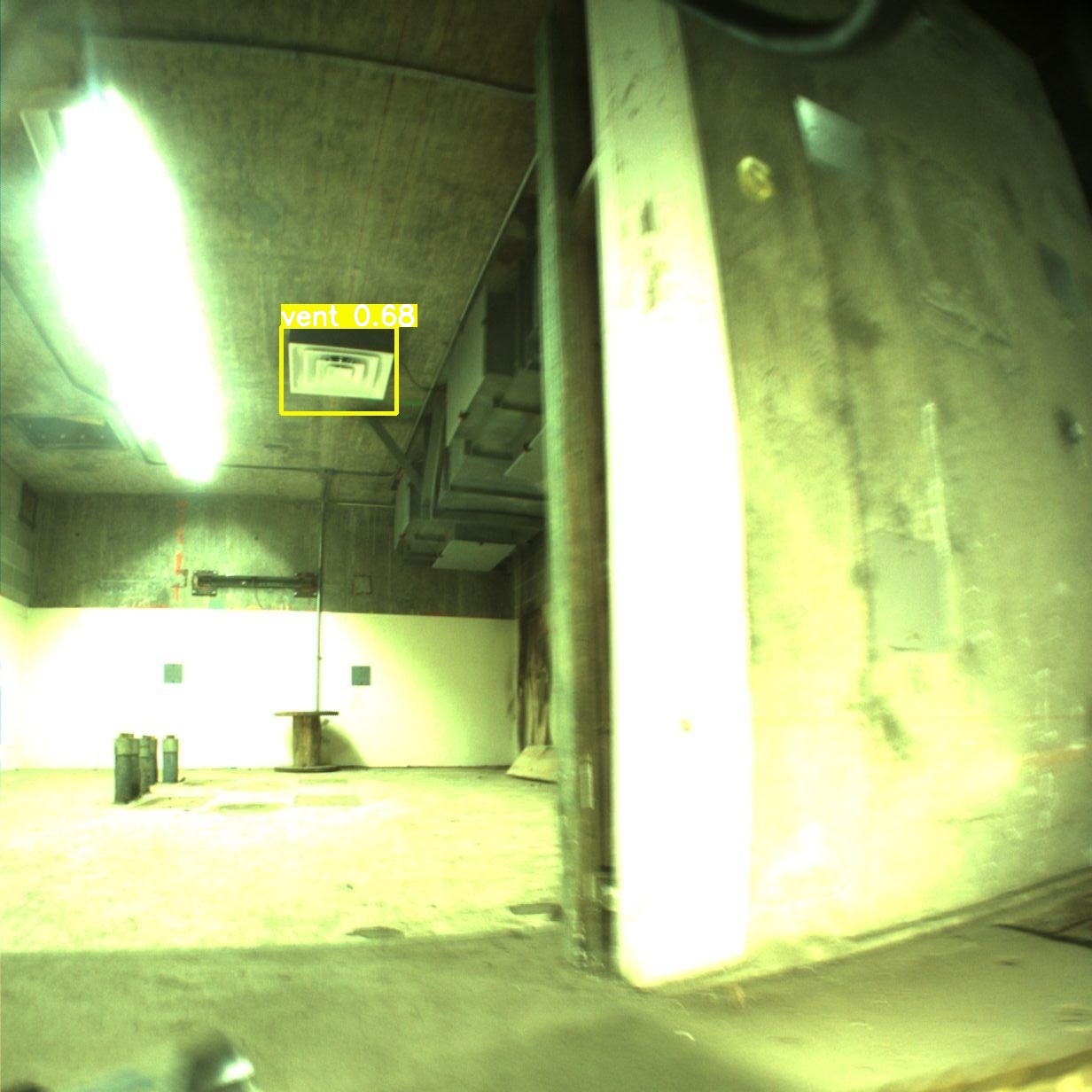}
	}
    \subfloat[False Backpack]{%
    	\includegraphics[width=0.24\textwidth]{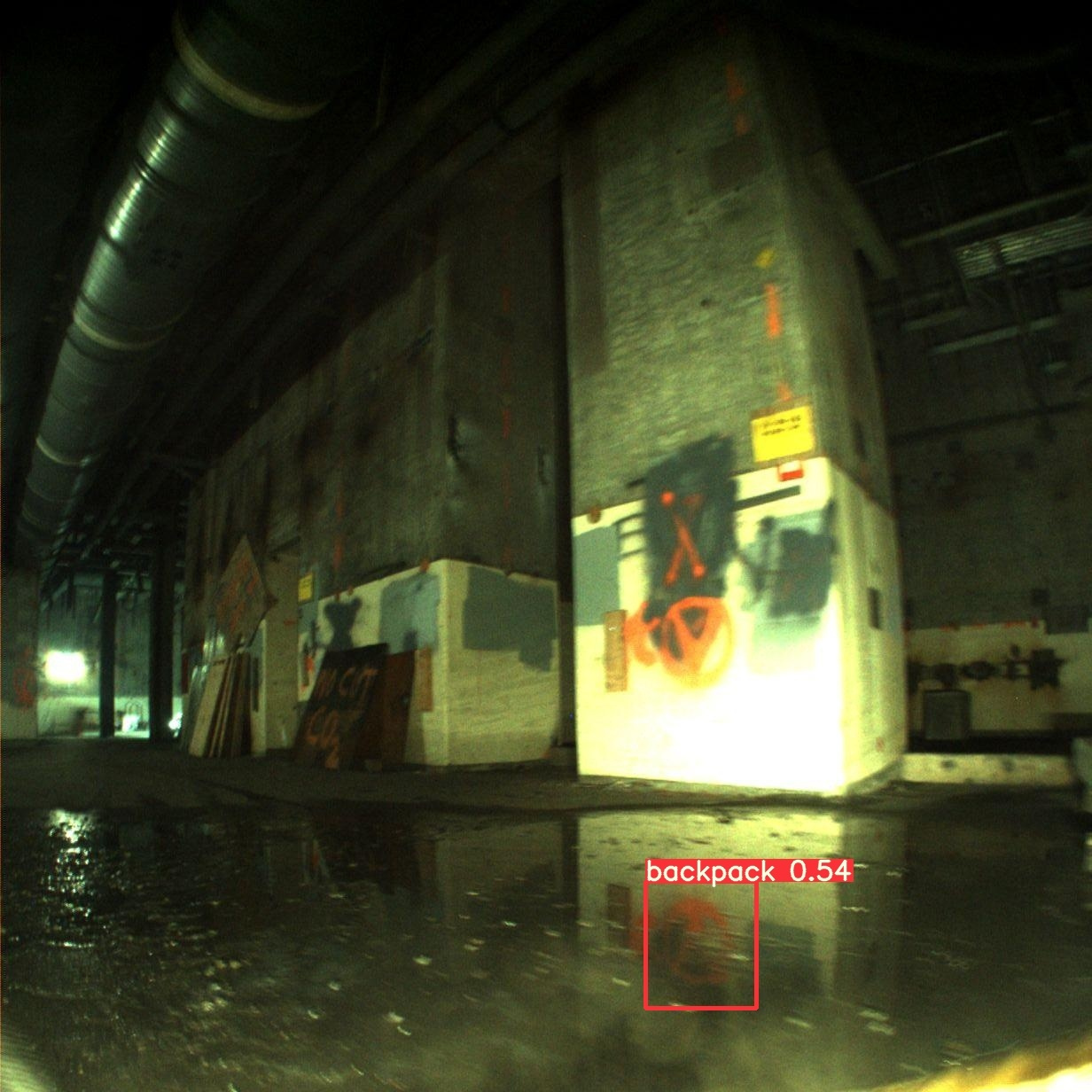}
    }
    \subfloat[False Cellphone]{%
    	\includegraphics[width=0.24\textwidth]{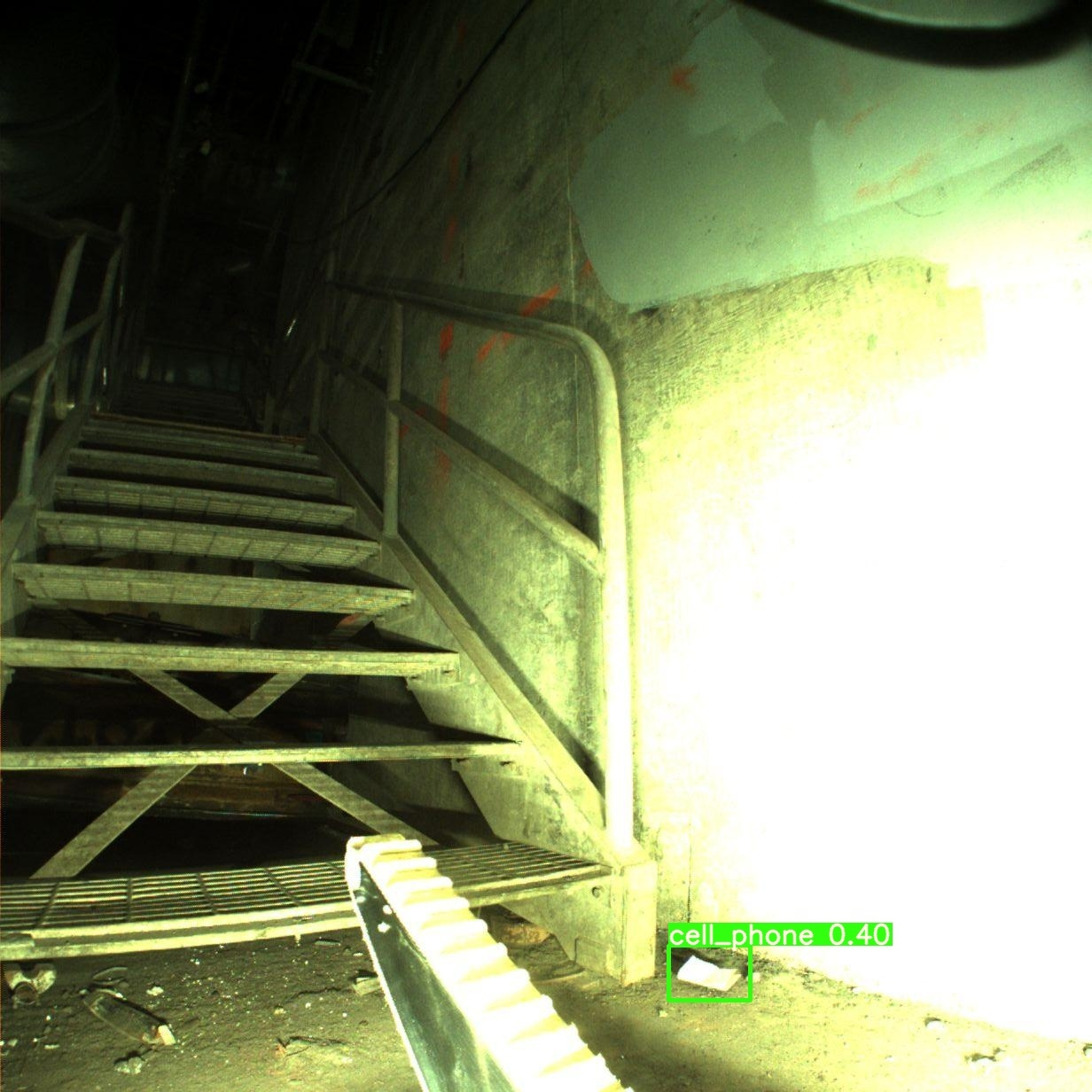}
    }
	\caption{\rnd{Detection of all object classes during the challenge. Two false positives are depicted, Backpack as the red sign in water reflection and cellphone as a piece of paper.}} 
	\label{pic:detections}
\end{figure}

If the robot has a 3D map of its surroundings, the bounding box provided by the YOLOv3 detector is projected into the 3D map. 
In case that the depth of the object in the image is not known, we save the vector direction between the camera and detected object and estimate the position later from multiple measurements using triangulation. 
Then, the detections are processed by the final position estimator \cite{vrba2020}, where each new detection creates a new hypothesis (described by state, position estimation covariance and the number of corresponding measurements) or is assigned to the already existing hypothesis. 
The final position of the hypotheses is established by the application of Kalman filtering.
Only if the hypothesis is certain (covariance lower then threshold and the satisfying number of measurements), the robot sends position, established class and RGB snapshot of the detected object to the database.
This prevents flooding of the communication network with image data and helps to lower the number of false positives detections send to the operator.
All the objects that were in the field of view of the robots during DARPA SubT scored runs were detected, and the number of false-positive detections was manageable \rnd{(Table \ref{tbl:urban_scoring})}, which indicates the good
performance of the detection method \rnd{(see Figure \ref{pic:detections} for the qualitative results).}

To ease the RGB detections and to improve the chance of finding phones, the Jetson TX2 was utilized to search surrounding for WiFi signal and using weighted trilateration estimate the position of the artefact~\cite{multilateration}.
Since in the Urban circuit, it was necessary to detect gas in the form of elevated CO2, we used Sensorion SDC30, which is sensor dedicated to measuring just CO2. Using sensor data, we extrapolated and
showed, not only vales to the operator but also eventual guesses where the CO2 artefact might be. \rnb{Both methods are explained in~\cite{tomavs2020senzoricka}.}

\subsection{Localisation and mapping}
\label{sec:localisation}
Detection of the artefacts is not enough. One has to determine their position with 5~m accuracy within a global coordinate frame provided by the contest organizer.
Therefore, after the object is detected, a robot has to determine the artefact position relatively to itself and then transform the position to global coordinates.
For that, a robot, which detects the object, has to know its position with sufficient accuracy.
Nowadays localisation methods report~\cite{vloam,orbslam,direct-slam} errors around 1\%, which, in theory, is sufficient for robots to venture more than 500~m deep into the underground environments before losing the ability to score any point by detecting objects.   
However, as discussed before, many of these tests are reported in ideal conditions only and operation in adverse conditions is investigated only rarely~\cite{santos2015sensor}. 
Our preliminary tests indicated that there is no universal solution that would suit all of the robots deployed by our team.
The problem of self-localisation, i.e., reliable estimation of the robot position, is tackled differently depending on the given platform.

\begin{figure}
	\subfloat[Alpha course, first run.\label{subfig-1}]{%
		\includegraphics[width=0.49\textwidth]{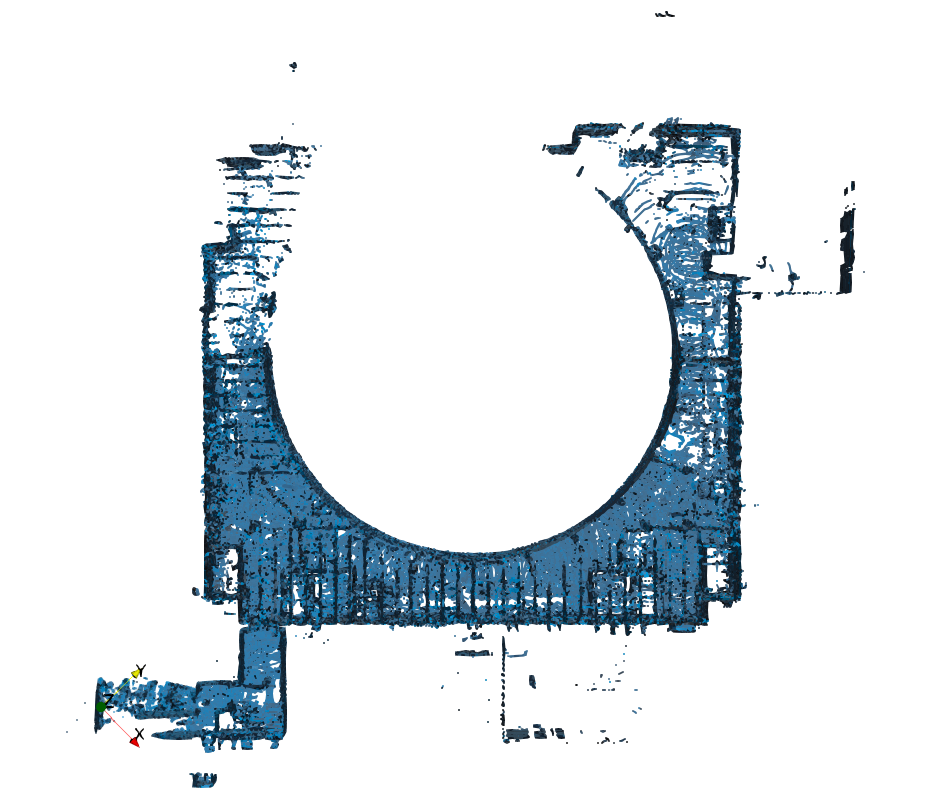}
	}
    \hfill
	\subfloat[Beta course, first run.\label{subfig-2}]{%
		\includegraphics[width=0.49\textwidth]{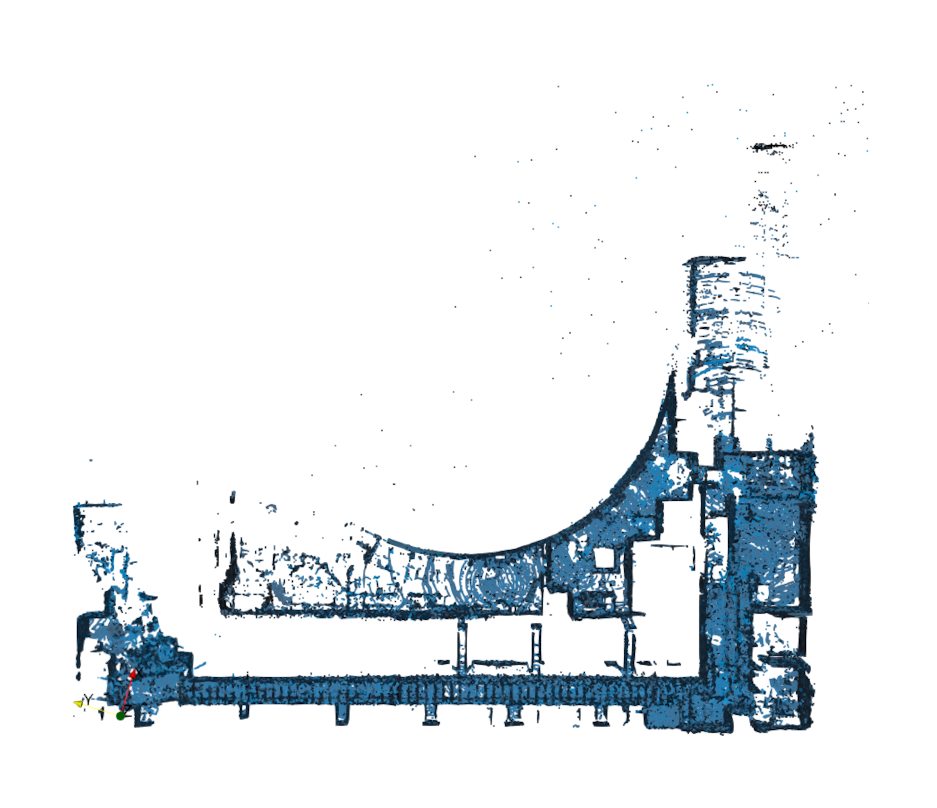}
	}\\
    \subfloat[Alpha course, second run.\label{subfig-3}]{%
    	\includegraphics[width=0.49\textwidth]{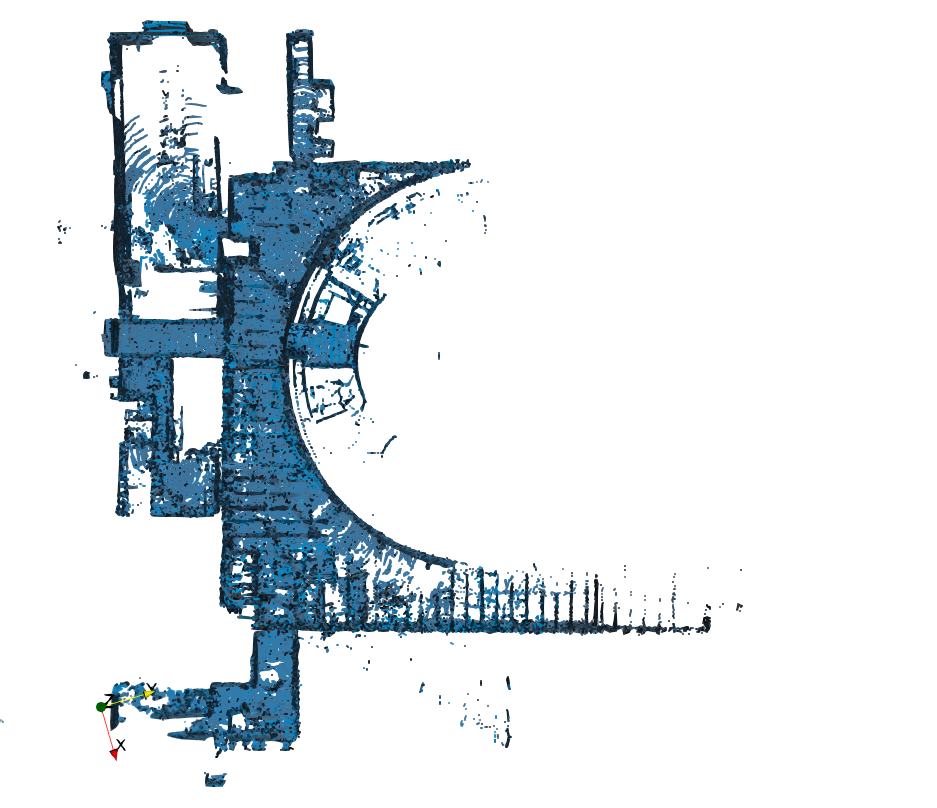}
    }
    \hfill
    \subfloat[Beta course, second run.\label{subfig-4}]{%
    	\includegraphics[width=0.49\textwidth]{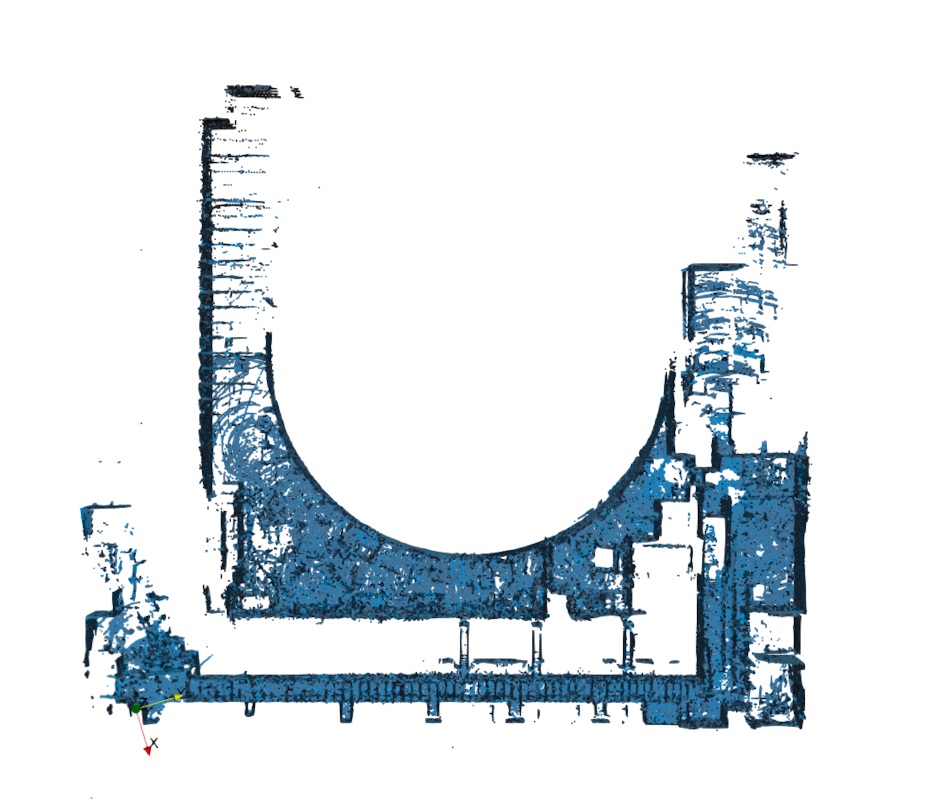}
    }\\
    \subfloat[Ground truth map of the Alpha course.\label{subfig-5}]{%
    	\includegraphics[width=0.49\textwidth]{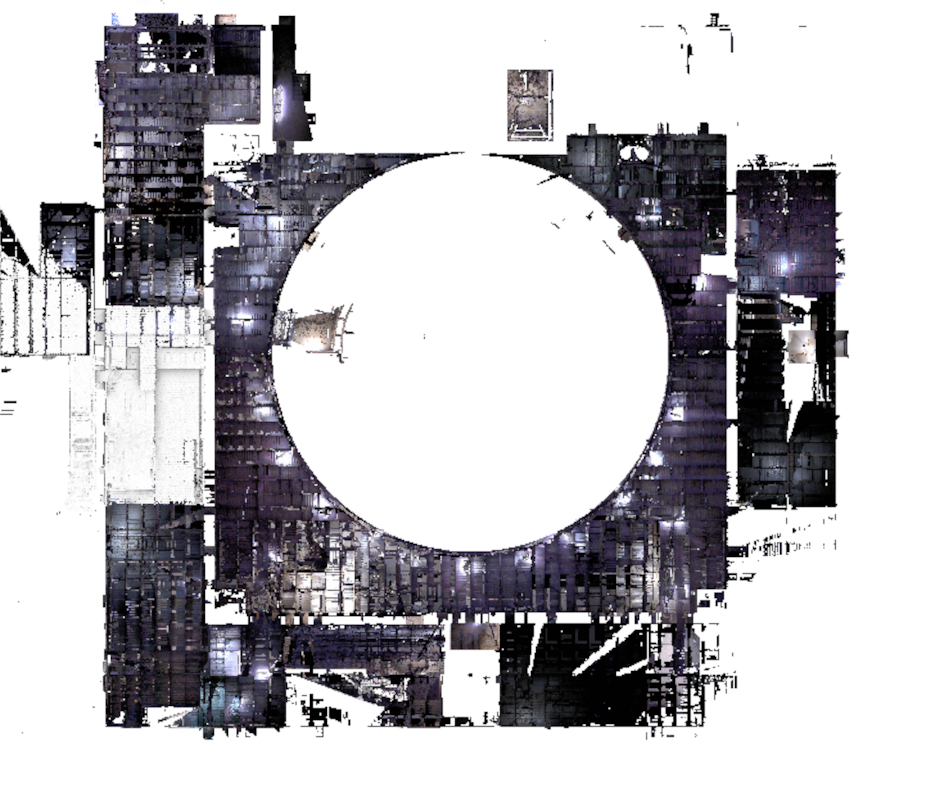} %
    }
    \hfill
    \subfloat[Ground truth map of the Beta course.\label{subfig-6}]{%
    	\includegraphics[width=0.49\textwidth]{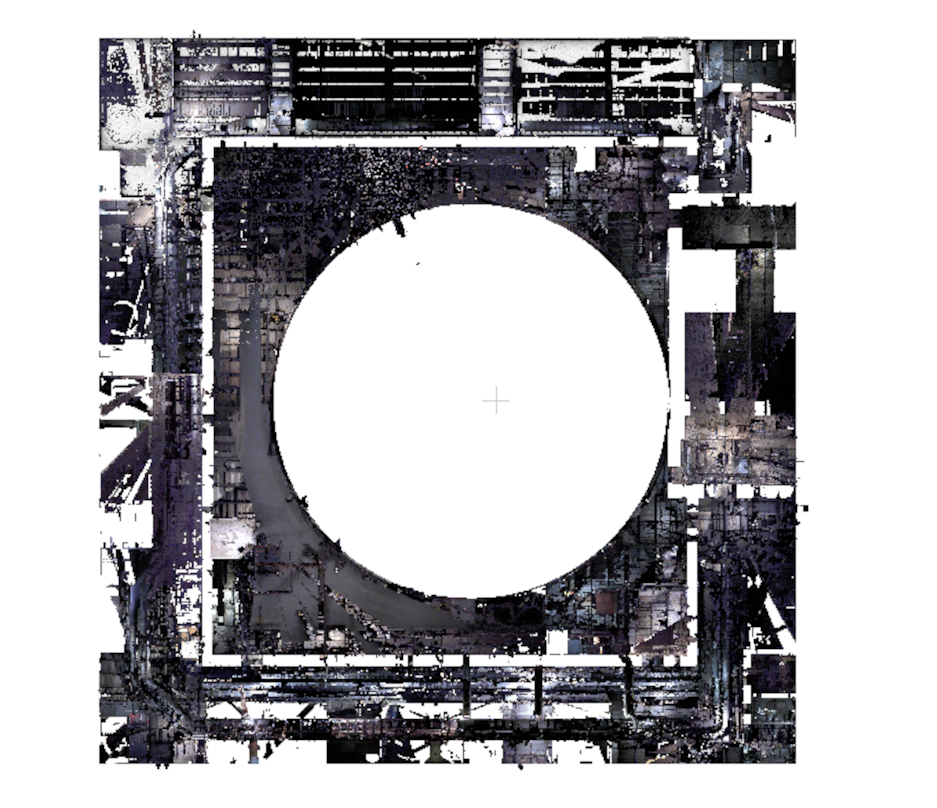}
    }
	\caption{Mapping results of the Husky robot (top-down view) from the Urban Circuit along with the ground truth maps for the Alpha (\ref{subfig-5}) and Beta (\ref{subfig-6}) courses.}
	\label{pic:maps}
\end{figure}

The wheeled and tracked robots employ EKF-based fusion and complementary filtering~\cite{Kubelka-2012-ICRA,Simanek-2015-TMECH} to provide dead-reckoning localisation and an initial guess to the simultaneous localisation and mapping (SLAM) method~\cite{Pomerleau2013,Pomerleau-2014-ICRA} based on the iterative closest point (ICP) method~\cite{Chen-1992-IVC}.
During SLAM, 3D scans from the lidar sensors are being aligned to an incrementally built global map, separately for each robot (see Figure~\ref{pic:maps} which presents the Urban Circuit maps from the Husky robot).
Since the maps created during the Tunnel Circuit showed bending effects in the initial long corridor, we applied gravity-vector-based constraints in the Urban Circuit.
They are an extension of the ideas presented in \cite{Babin2019a} and they constrain the ICP algorithm in the roll and pitch angles.
Accuracy of dead-reckoning localisation of the tracked robots, which suffers when traversing vertical obstacles, is improved using a combination of explicit modelling of robot kinematics and a learned support-vector machine (SVM) model~\cite{Kubelka-2019-TMECH}.

\begin{figure}
	\centering
	\includegraphics[width=0.8\textwidth]{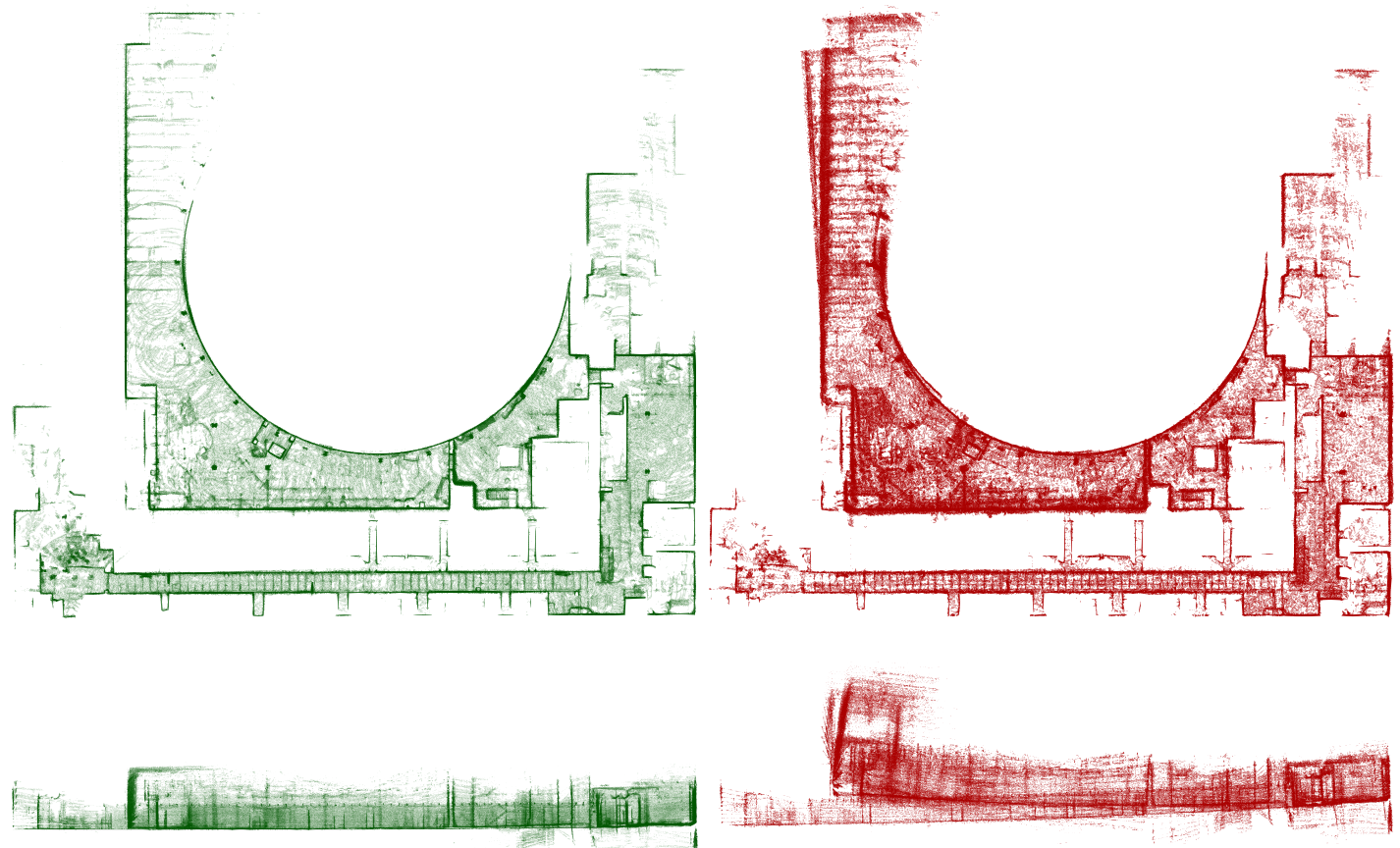}
	\caption{\rnd{Maps of the Beta course obtained from our localisation pipeline (left, green) and from the A-LOAM algorithm (right, red). The sensor data come from the Husky robot during the second Beta run. The top row presents a top-down view, the bottom shows side-views of the resulting maps.}}
	\label{pic:loc_compare}
\end{figure}

\rna{By a design choice, we do not perform loop closure.
Instead, the localisation pipeline is tuned to stay within the required accuracy to successfully report the detected artefacts.
In the ICP algorithm, the chosen cost function for matching point clouds is the distance between the lidar points and the map planes (\emph{point-to-plane} minimizer).
We have found this to be robust for the environments we observe in the SubT challenge.
The main concern is the localization drift which manifests itself in map bending and warping.}
\rnd{To demonstrate the character of mapped environment, we ran the A-LOAM\footnote{\url{https://github.com/HKUST-Aerial-Robotics/A-LOAM}} algorithm, an advanced implementation of the original Lidar Odometry and Mapping (LOAM)~\cite{zhang2014loam}, on data recorded by our Husky robot in the second Alpha and Beta sorties.
Figure~\ref{pic:loc_compare} shows the A-LOAM and ICP maps of the Beta sortie.
The robot entered through the long south corridor and followed the circular reactor wall.
In this situation when loop closure is not possible, it is hard to avoid localization drift.
The drift can be only reduced, not eliminated, by tuning the algorithm, by adding complementary measurements or by profiting from some a priori knowledge (e.g. that the floors are usually level).
An example of a complementary measurement for a LOAM-like localization is the work of Ye~\cite{Ye2019}.
In our approach, we can constrain directly the ICP algorithm to align the point clouds with the measured gravity vector.}

\begin{figure}
	\centering
	\subfloat[Beta course, second run. Distance driven: $260\,$m \label{subfig_loc-1}]{%
		\includegraphics[width=0.9\textwidth]{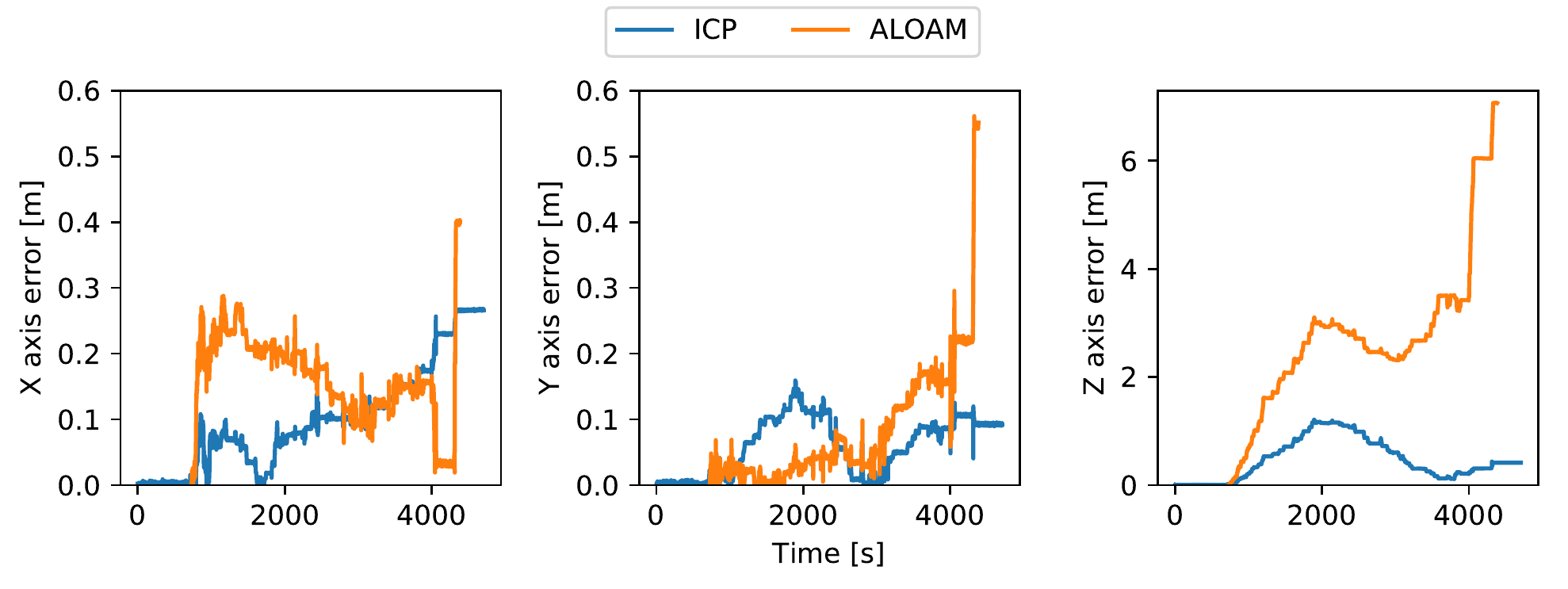}
	}\\
	\subfloat[Alpha course, second run. Distance driven: $247\,$m \label{subfig_loc-2}]{%
		\includegraphics[width=0.9\textwidth]{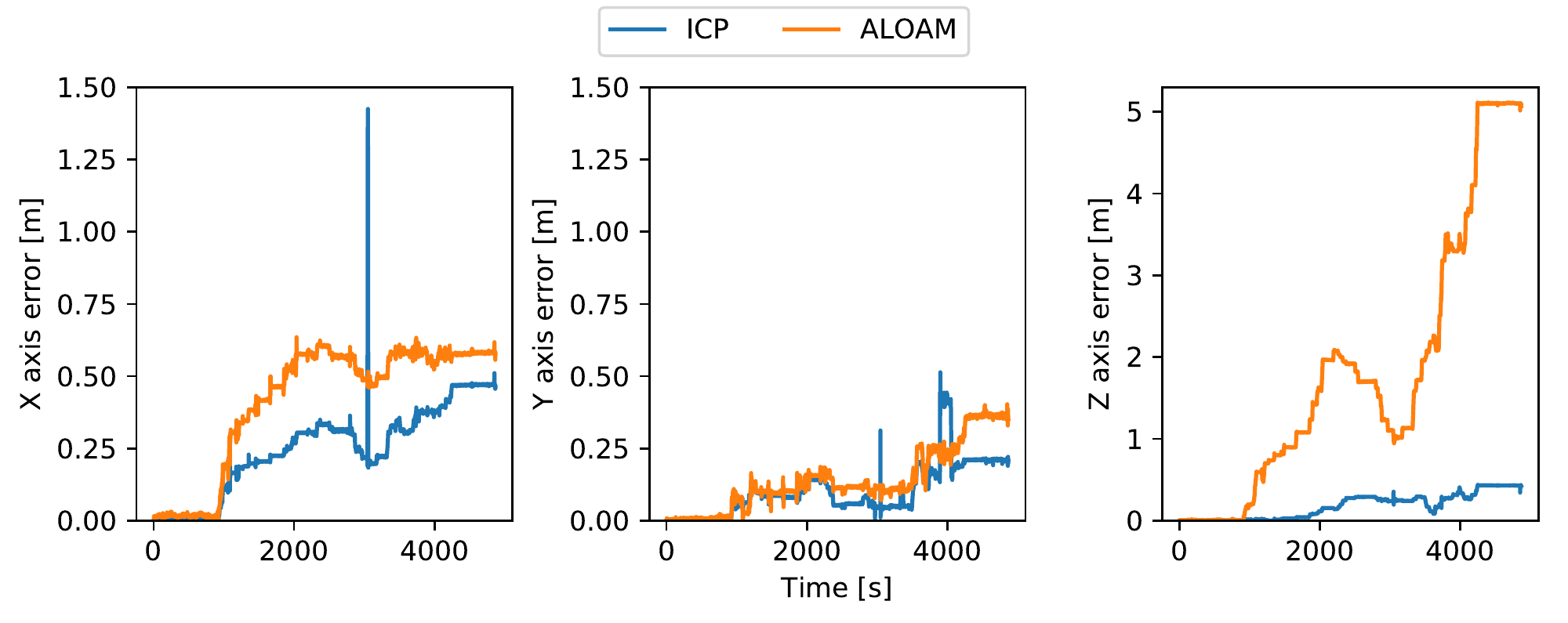}
	}
	\caption{\rnd{Localization error of our ICP-based solution and of the A-LOAM algorithm. The error is prominent the most in the Z axis, note the different scale in the third column.}}
	\label{pic:loc_drift}
\end{figure}

\rnd{Moreover, thanks to ground-truth position computed by localizing in the maps provided by DARPA after the Urban circuit, we are able to evaluate localization error.
The plot in Figure~\ref{pic:loc_drift} demonstrates that the main problem is the drift in the Z-axis caused by the map bending.
In the case of the Beta course, the highest error around the time $2000\,$s corresponds to relative error lower than 1\%.
This result supports our decision not to strive for loop-closure mechanisms in this application.
However, monolithic point-cloud maps also yield some disadvantages.
Computational complexity grows with space explored and mapped, but this can be solved by appropriate memory management.
A bigger problem that we encountered were erroneously aligned point clouds that damaged the map.
These misalignments usually occurred when navigation over rubble and introducing high rotational speeds, or when crashing into obstacles.
To cope with this problem, we are currently investigating available point-cloud motion-deskewing methods.
Finally, there were several instances where the rotating 2D lidar on the tracked robots would stop working, mainly because of crashing into obstacles and stopping the servo drive.
This mechanical liability will be solved replacing the 2D lidars with modern multi-beam, 360-degree ones.}

The UAVs were initially localised by a Kalman-filtering-based state estimation system \cite{petrlik2020ral}, which was fusing measurements from IMU with position corrections obtained from Hector SLAM~\cite{hector}.
Hector SLAM is a motion estimation algorithm explicitly developed for search and rescue (SAR) scenarios.
The 2D laser scans generated by the RPLidar A3 laser scanner are first transformed into a local stabilised coordinate system using the UAV orientation measured by IMU.
After preprocessing the laser scan by discarding the endpoints that could potentially originate from the ground or the ceiling, the rest of the points are aligned into the online-estimated occupancy grid map.

The localisation method of the UAVs was changed after Tunnel circuit due to a more vertically diverse environment of the Urban circuit.
The assumption of similar cross-section among different heights that was used in the Tunnel circuit was no longer valid.
To be able to navigate the more complex building interiors, the 2D lidar was replaced by the 3D (multi-plane) lidar Ouster 0S1-16 with 16 measurement planes.
Since Hector SLAM was not developed for full 6-DOF pose estimation, the position corrections for the UAV state estimation are provided by the A-LOAM.
A-LOAM is based on extracting edge and planar features from the input point cloud, which are then matched into the global map represented by a gradually built point cloud.
\rnd{Contrarily to the localization pipeline run in the UGVs, A-LOAM does not require prior motion estimates from odometry and it is therefore a more suitable choice.}
To assist the localisation and mapping pipeline, an adaptive intensity-based filter removes irrelevant data from the Ouster OS1 lidar in order to filter out clouds of whirled dust, which emerged in both the circuits due to the aerodynamic influence of the UAVs.

The localisation of the hexapods is based on the Intel T265 camera module which provides a position estimate based on a proprietary grey-scale visual-inertial SLAM algorithm.
The robots are equipped with a secondary RGBD camera (Intel D435) which builds a 3D map~\cite{bayer19ecmr}, and uses the map to guide the robot to unexplored areas.  
\subsection{Navigation}
\label{sec:navigation}
\subsubsection{UGV}
Besides the map for the localization, each ground robot builds a separate map for navigation and exploration purposes. 
The navigation map is built from sensors available on the particular robot; walking robots used depth camera Intel RealSense D435, tracked robots and UAVs used 2D LIDAR, and wheeled robot fused range measurements from the 3D lidar with two depth cameras Intel RealSense D435. 
More in depth priciples of navigational maps are based on published in work~\cite{bayer19ecmr}. 
The map representation for walking robots used the elevation map with the underlying quad-tree structures, and cache enhancement described in~\cite{bayer20mesas}. 
For tracked robots and the wheeled robot, the map representation has further been extended to 3D by applying a custom speeded-up oct-tree data structure. 
The navigation maps were periodically analyzed for terrain traversability to determine how to navigate the robot to the specified location. 
The traversability is estimated from the roughness of the terrain. 
When the roughness exceeds threshold $k_d$, the terrain is considered untraversable.
Where $k_d$ has been estimated for each robot type based on the robot kinematics and experimental results. 
In the opposite case, the terrain is traversable, and a path generated by the planner can be planned over the corresponding map cell. 
Plans for the robot position in the map to the single given goal location were produced by the A* algorithm with Euclidean distance heuristic.
Moreover, the cost field generated using a generalized version of the Distance transform algorithm~\cite{felzenszwalb12dt} for 3D maps have been utilized by the planner to keep a safe distance from the obstacles or to avoid rough but traversable terrain if possible. 
We have dedicated the extra module for the following of the generated path. 
The module itself smoothes the path using a sliding window filter, and the combination of the two control laws ensures that the robot follows the path. 
The first control law steers the robot based on the displacement of the robot orientation, and the second control law sets the forward velocity, based on the distance to the closest obstacle, and distance to the next navigational waypoint.

The motion planning method used proved to be universally applicable, and it works well on our crawling, wheeled and tracked robots.
\rna{
\subsubsection{Adaptive terrain traversal}
To improve their ability to overcome adverse terrain, the tracked robots incorporate the information from their RGBD cameras and position their auxiliary tracks accordingly~\cite{flippers1,flippers2}.
Thus, flipper control was done autonomously most of the time.
This is because manual control would cause excessive cognitive load on the operator, especially when operating several platforms simultaneously.
Manual control is also time-consuming as there is a several second delay between issuing the commands and receiving visual feedback of the flipper position.
The flipper control algorithm is reactive and uses information from the IMU and a depth input from a front-mounted D435 camera.
}
\begin{figure}[h]
  \begin{center}
  \includegraphics[width=6cm]{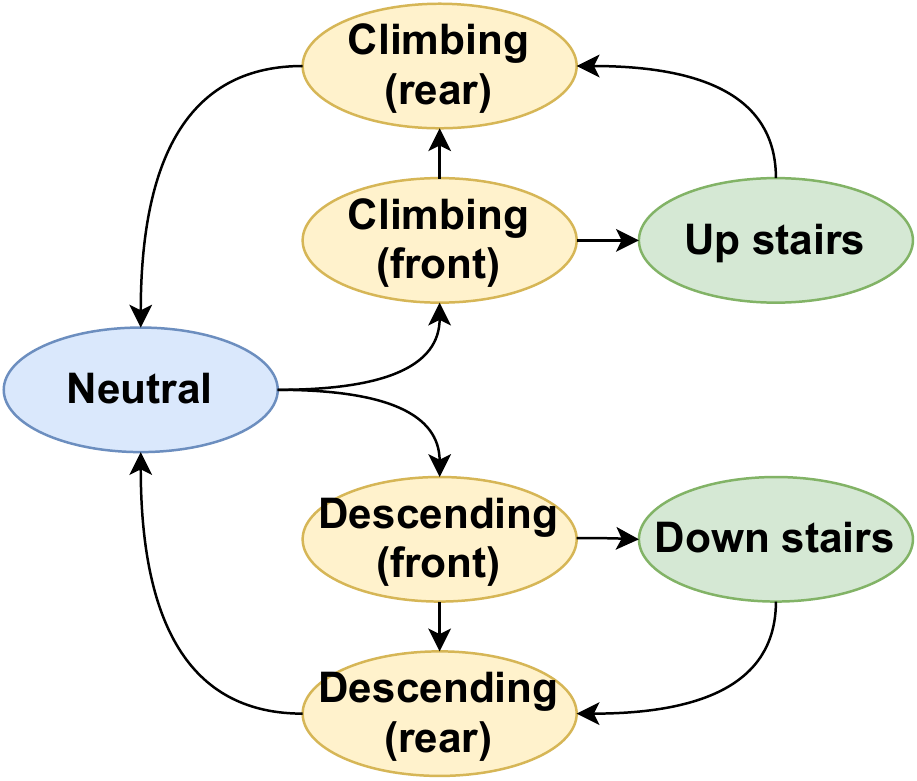}
  \caption{\rna{Finite state machine for controlling flippers of the Absolem platform, capable of traversing difficult obstacles and stairs. Blue and green states signify longer lasting modes, whereas yellow states are short lasting.}}
  \label{fig:fsm}
  \end{center}
\end{figure}
\rna{
Our approach consists of a state machine, shown in figure \ref{fig:fsm}, where we define four primary states of obstacle traversability.
The neutral state means folded flippers in such a way to minimize blocking the Lidar.
Yellow states define configurations of the flippers which allow the robot to safely climb and descend obstacles by using the front and back flippers as support.
These are triggered when the robot is moving forward on the flat ground, and a significant height increase or decrease is sensed in front of the robot in a designated region.
For example, in the \textit{Climbing rear} state the rear flippers push down to raise the rear while keeping the front flippers extended forward to shift the weight balance maximally forward and dampen the robot when it hinges forward.
Whenever the robot finishes the ascent or descent on to flat ground the state resets back into neutral.
The same state machine is used for stair traversal with a modification that the robot cannot turn while on a stairs.
When setting flippers for any action, we make use of the natural tactile feedback of the ground by simply pressing down with low torque.
This is better, than strictly conforming the flippers to a height map which can be prone to errors.
}

\subsubsection{UAV}
As the aerial robots do not move on the ground, they do not need to consider the traversability of the terrain.
Thanks to that, the UAV can potentially explore areas unreachable by the ground robots.
The areas accessible by the UAVs are, however, limited by the narrowness of the corridors leading to these areas. 
With decreasing width of the narrow passage, the probability of a collision that would render the UAV incapable of continuing the mission rises due to inaccuracies in the localization and wind gusts caused by the propellers in tight areas.
To achieve high reliability while simultaneously avoiding deadlocks caused by narrow passages, an exploration method that favours paths further from obstacles was developed. 

For the Tunnel circuit, the navigation is based on the 2D occupancy grid produced by Hector SLAM.
First, the grid is downsampled, the obstacles are inflated, and then the distance of each cell to the closest obstacle is obtained by a distance transform.
A modified A$^*$ algorithm is run at \SI{2}{\hertz} to generate the path from the current UAV pose to the goal using the heuristic:
\begin{equation}
  h(\mathbf{x})=dist(\mathbf{x}, \mathbf{g}) + o({\mathbf{x})p},
\end{equation}
where $\mathbf{x}$ is the current evaluated point in the grid, $\mathbf{g}$ is the goal, $dist(\mathbf{x}, \mathbf{g})$ is the Euclidean distance, $o(\mathbf{x})$ is the distance to the closest obstacle, obtained from the distance transform, and $p$ is a tunable parameter, which controls how much the planning algorithm avoids obstacles.
The used heuristic forces the generated paths to keep a distance margin from obstacles and pass through narrow passages only when another path cannot be found.
The operator sets the goal, and with prior knowledge of the tunnel, the layout could be used to bias the exploration in a specific direction.
During the competition, the goal was set to \SI{300}{\metre} in the x-axis of the SubT reference frame to explore deep into the mine without any bias in the y-axis.
Depending on the chosen exploration strategy, the UAV can search for artefacts its whole flight time and relay the positions of artefacts over the mesh network formed by other robots.
The other option is to return to the initial position after reaching half of the flight time by setting the initial position as the new goal.

Due to the complexity of expected environments, the same approach could not be used for the Urban circuit.
Therefore for complex 3D environments, the UAVs identify the next goal from the set of exploration waypoints. 
Paths to the individual locations are generated with the A$^*$ algorithm with Manhattan distance heuristic applied on the volumetric map produced by the ALOAM mapping.
To overcome problems of trajectory-tracking control errors and possible dynamic obstacles, the planning process is sped up by utilizing only the 6-neighborhood and adaptive map resolution, and the current A$^*$ path is frequently replanned.
The initial or a replanned path is then post-processed with a fast iterative algorithm which smooths the path and increases an obstacle margin, providing safe navigation in a complex dynamic 3D environment in real-time.   

\subsection{Exploration}%
\label{sec:exploration}
Taking into account two circuits of the DARPA Subterranean competition, the two different exploration strategies were used for the ground robots.
The first strategy is the frontier based exploration~\cite{frontier}, which navigates the robot to the closest edge between free space and unknown space. 
We have employed the improved frontier based exploration strategy described in~\cite{bayer19ecmr}, which proposes to lower the computational demands of the next navigational goal selection from the possible goal candidates by frontier cell clustering.

The disadvantage of the frontier based exploration is that the environment coverage by the spatial map not corresponds with covering the environment by the robot cameras to search for the artefacts, especially when the robot is capable of detecting obstacles from greater distances than to recognize the artefacts.
Therefore, we have also introduced the exploration strategy that used the model of the robot cameras to generate such exploration waypoints, so the robot covers the environment by its cameras. 
The navigational waypoint, followed by the robot, was selected from the exploration waypoints based on the entropy, which estimates the information that the robot obtains by observing a particular location.

Plans to the close goals during the exploration were generated by local A* planner, or in case of multiple goal candidates by the Dijkstra algorithm. 
On the other hand, the homing that often required a very long plan took advantage of the already known map covered by the sparse graph connecting the selected traversable waypoints similar to the approach proposed in~\cite{dang19icar}. 
The human operator did coordination between the robots since he had complete information about maps of all robots.

To utilize the capabilities of UAVs, their exploration strategy aims at the exploration of further areas of the environment instead of a thorough exploration of nearby locations. 
In contrast to the ground robots, they utilize the map generated from the 3D lidar also for the generation of exploration waypoints.
A navigational waypoint for the UAV is chosen according to a custom cost function, which reflects the deviation from a specified direction, and horizontal and vertical distance which need to be travelled in order to reach the waypoint.

\subsection{Coordination}

Coordination of the robotic team is performed primarily through the powerful data-link provided by the 4G Mobilicom mesh transceivers.
This allows the operator to operate the robots directly or to set the goals where they should navigate.
Moreover, the robots are aware of each other's positions using a low-bandwidth link established by the droppable ``Motes'' described in subsection~\ref{sec:motes}. 
By remembering the shared positions, each robot can reconstruct the trajectory of each other robot in the team and avoid exploration of already-visited areas.
History of the positions also allows to detect situations where a robot is mislocalised or stuck.
\subsection{Interface}
\label{sec:interface}
\begin{figure}[!htb]
  \includegraphics[width=\columnwidth]{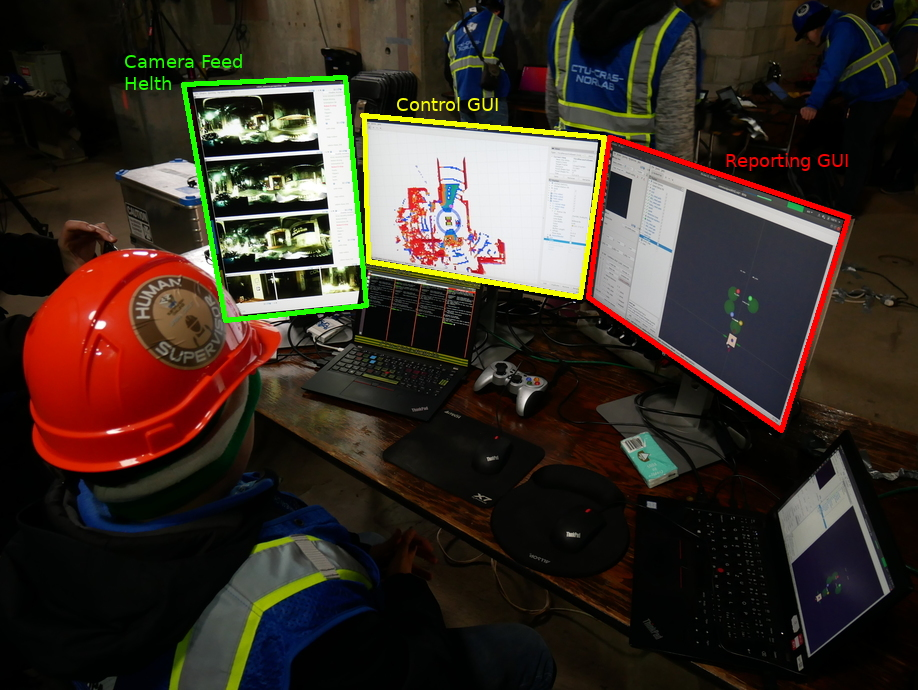}
  \caption{The base station and its user interface used in the Urban circuit event.\label{fig:base_station}}
\end{figure}

As the challenge rules allow only a single human supervisor to manage all the systems deployed in the mission, it was necessary to design a reliable and easy-to-use user interface (UI) to supervise and control the robot's behaviour and report back found artefacts.
With an increasing number of deployed systems, the UI plays a crucial role in the effectivity of the whole mission; therefore it is a part of the system that have undergone substantial modifications since the STIX event to increase its reliability, to cope with the modifications in the autonomous behaviours of the robots, and to simplify the work of the human supervisor.
The current UI is composed of two major parts including an interface for robot control and interface for artefact reporting and several diagnostics utilities that allow the pit-crew to visualize the telemetry data from the robots and simplify the deployment process of the robots during the preparation phase before the scored run.
The two major parts of the UI that are used to control the robots and report the artefacts are realized in the base station that is visualized in Figure~\ref{fig:base_station}. %

\begin{figure}[!htb]
\begin{center}
    \includegraphics[width=1.0\columnwidth]{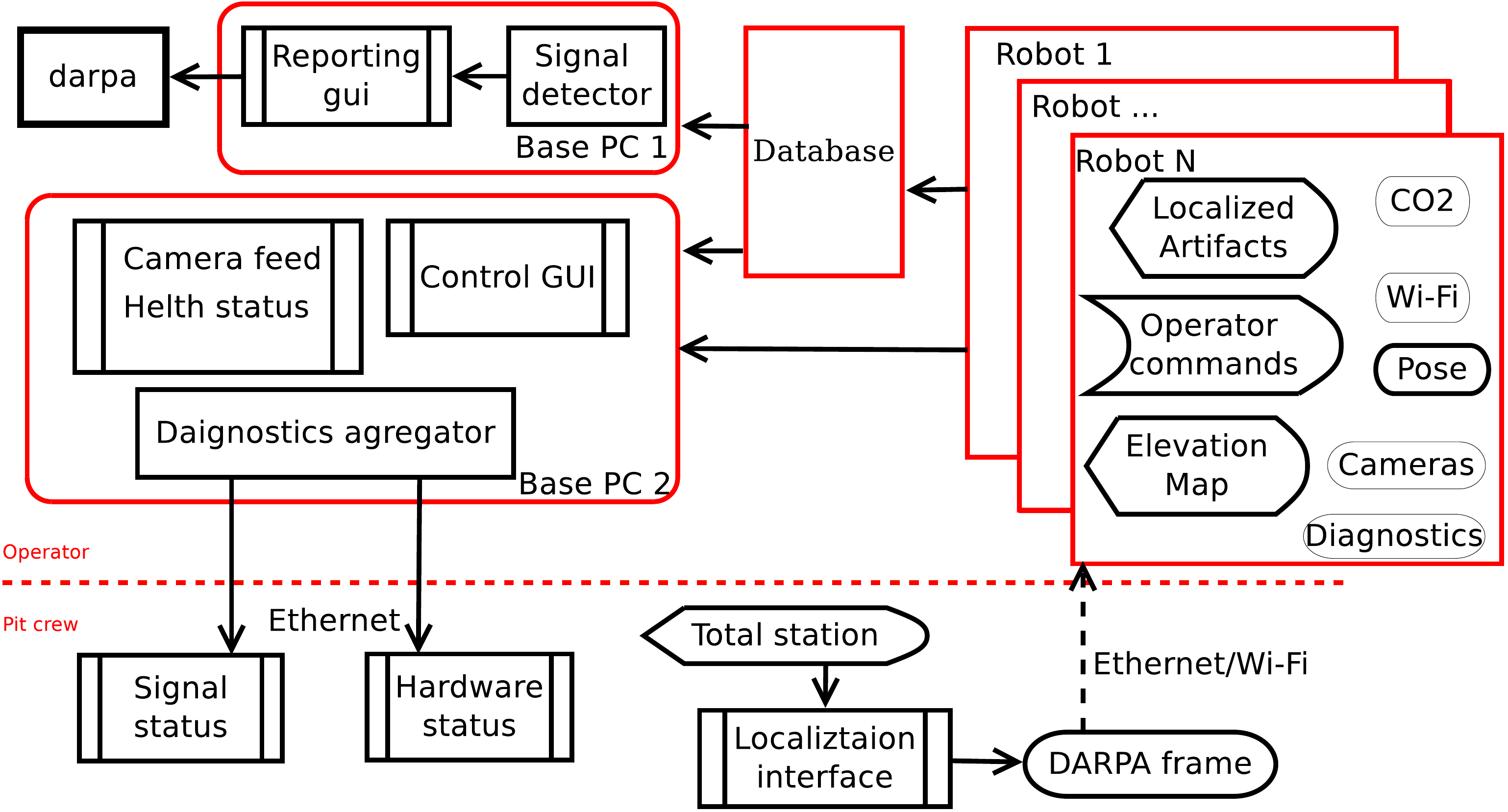}
    \caption{\rna{Block/ diagram of base station modules with status monitors for pit crew.}} 
\end{center}
\end{figure}

The base station UI is built on top of standard ROS visualization components, namely custom rqt plugins and rviz tool.
The robot control UI is divided into two screens.
The first screen shows the video feed and current status (e.g. localization status, battery level, emergency stop state) from individual robots using customized rqt plugins fitted in a single non-interactive rqt perspective.
The second screen of the robot control UI uses interactive marker server and multiple custom plugins, e.g., to capture keyboard events, in rviz for control of individual robots.
It allows to show and hide the constructed maps of individual robots and their horizontal cross-sections to deliver the human supervisor the spatial awareness, and assign high-level commands like follow the target or explore using an interactive marker and keyboard shortcuts.
Note that so far only the four large robots are controlled using this interface, the drones and hexapods are entirely autonomous, and they interact only with the artefact reporting UI.

The artefact reporting UI serves the purpose of accumulating the detections from all the robots and presenting these detections to the human supervisor that verifies them and sends them to the DARPA scoring server.
The UI is realized as a custom rqt plugin and a ROS node that visualizes the individual detections using the interactive marker server in rviz.  
As the detections are transmitted to the base station with their image, the rqt plugin presents this image to the human supervisor.
The human supervisor may iterate through the detections, confirm or reject them, change any parameter of the detection, and finally he may send the detection to the DARPA scoring server.
On top of showing the human supervisor the detection images in the rqt plugin, the detections are visualized inside the rviz as manipulatable interactive markers.
The rviz visualizes the individual robots, their path, the detections, and also the strength of the Wi-Fi signal emitted by the cell-phone artefact and the CO$_2$ levels essential to detect the gas artefact that is measured by the robots along their path.
The wi-fi and CO$_2$ levels are visualized as colour-coded point clouds along the robot path.
The rqt plugin allows switching the visualization of the individual elements for each of the robots in rviz on and off.
Further, the detections visualized as interactive markers are colour- and shape-coded based on their status, e.g., detections confirmed and rejected by DARPA are green and red spheres respectively, detections rejected by the human supervisor are small grey cubes. 
This allows the human supervisor to recognize the state of the detections immediately. 
Last but not least, the human supervisor is able to manipulate the spatial positions of the detections, and thus manually correct for the localization drift of the robots prior to sending the artefact positions to the DARPA scoring server. 
This feature has proved to be essential in reporting of many artefacts during 'Tunnel' circuit where the robots had significant map drift. 
\section{Experiments and deployments}

Before the ``Urban'' circuit at the discontinued nuclear power plant ``Satsop'' our team deployed the robotic system on four other different occasions: ``Edgar'' -- Coal Mine in Idaho Springs, ``URC
Josef'' -- a CTU experimental mine, ``NIOSH'' -- mine facility near Pittsburgh and undisclosed location inside Prague ``subway'' infrastructure. \rnd{ Additional deployment in the ``Bull rock'' cave was
performed as a preparation for the ``cave'' circuit which eventually did not take place due to the global pandemic. Our team participated in the virtual track after it had been announced that the systems track is
canceled.  Several other small-scale experiments were conducted mostly for testing individual parts of a system.
Gathered data, including everything collected during our runs and other deployments, is currently available online\footnote{\url{https://login.rci.cvut.cz/data/darpa-subt/data/}}.
This dataset is still not complete as new data is still being added and will be structured in the future after the whole competition is over.}

\subsection{STIX}
\begin{figure}[!htb]
  \begin{center}
    \includegraphics[width=1.0\columnwidth]{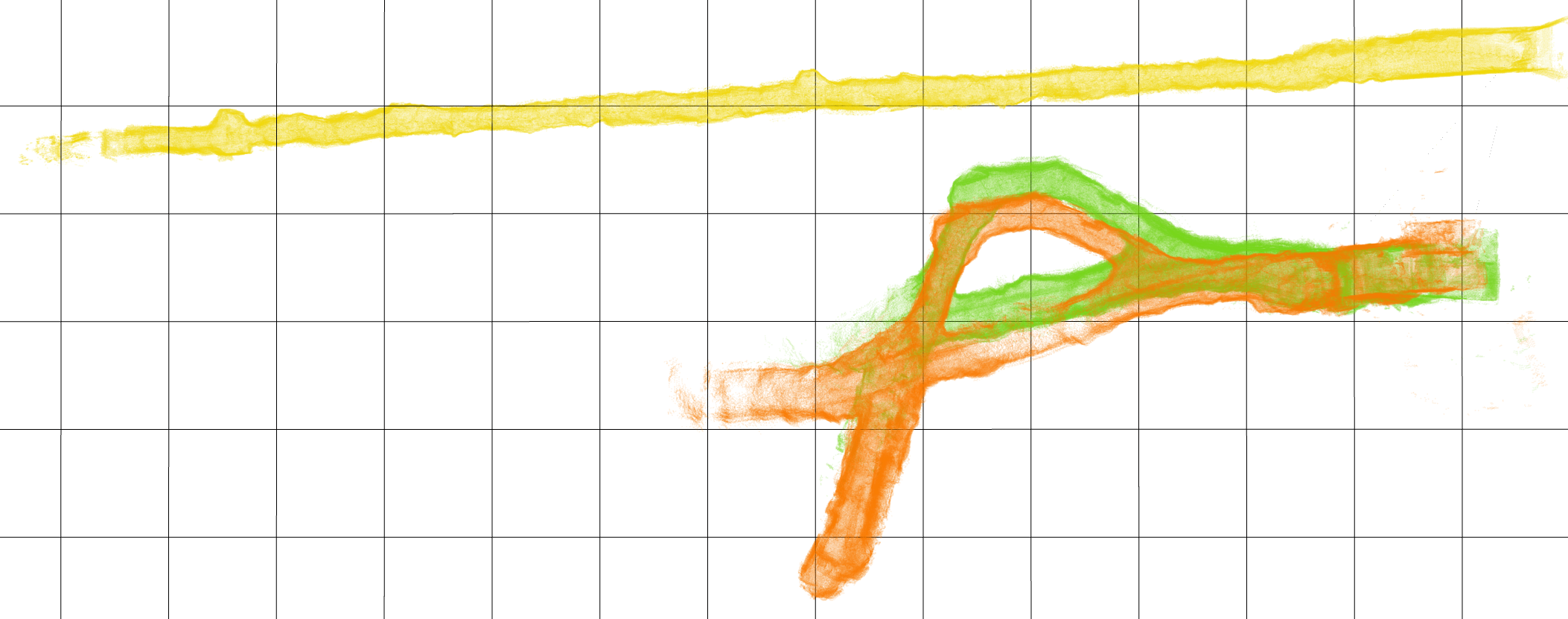}
    \caption{Map of  ``Miami''~(top) and ``ARMY''~(bottom) courses during STIX event. Notice improper aligment of an angle to the DARPA frame in the ARMY tunnel. 10m grid.} 
\label{fig:stix}
\end{center}
\end{figure}

The first deployment took place at the ``Edgar'' mine, where DARPA organized the STIX event. %
This event focused on testing the teams' readiness and DARPA's infrastructure for later competition circuits.
In this deployment, four runs, each lasting three hours, were performed \rnd{on two ``ARMY '' and ``Miami '' tracks.}
During these runs, the robots could detect and correctly localize three objects\rnd{, even though they were controlled only by operator joystick.
UAVs were deployed only on ``Miami '' to explore the first 50 meters autonomously before crashing due to a narrow corridor of the tunnel.
Track  ``ARMY '' was not favorable for drones due to the enormous amount of dust that hindered all platforms when the UAVs tried to lift off.
No proper alignment to the DARPA map was implemented, and we have relied on starting the robots mapping on a specific place in front of the gate to the course.
This resulted in improperly aligned maps of each robot as seen in~\ref{fig:stix}.
}
Several problems had arisen during those tests, such as communication issues between robots caused by improper Mobilicom network settings and hardware failures\rnd{, and issues with overseas battery transport.}
This lead to a need for faster robots (Husky), hardware improvements of other platforms, improved operator UI, and more frequent in-house testing.   
After all four runs were completed, the DARPA representative initiated a tour throughout the mines to demonstrate the scale and hazards that our robots will face in subsequent circuits.
Those hazards included an area obscured by fog generator, constrained passages, rubble, and masterfully hidden artefacts visible only from specific angles or beneath the metal grating floor.

\subsection{Josef}

Our team held the second deployment of the robotic team in the ``Josef'' experimental mine managed by the Czech Technical University in Prague.\rnd{
The team `Robotika'' also utilized this site simultaneously since it was one of few options of such tunnels accessible in the Czech Republic.}
This deployment consisted of several days in an actual mine to test and evaluate the system's features.
Deployment at the ``Josef'' mine enabled us to gather more specific datasets for better evaluation of localization and mapping and training data for the neural network for object detection.
We also gained new experience on robot behavior when driving them over gravel, wet railways, or through a light haze.
Additionally, we realized the need to implement specialized modules, such as water and wet surface detection, to tackle the unreliability of rangefinding sensors scanning those types of materials.
\begin{figure}[!htb]
\begin{center}
    \includegraphics[width=1.0\columnwidth]{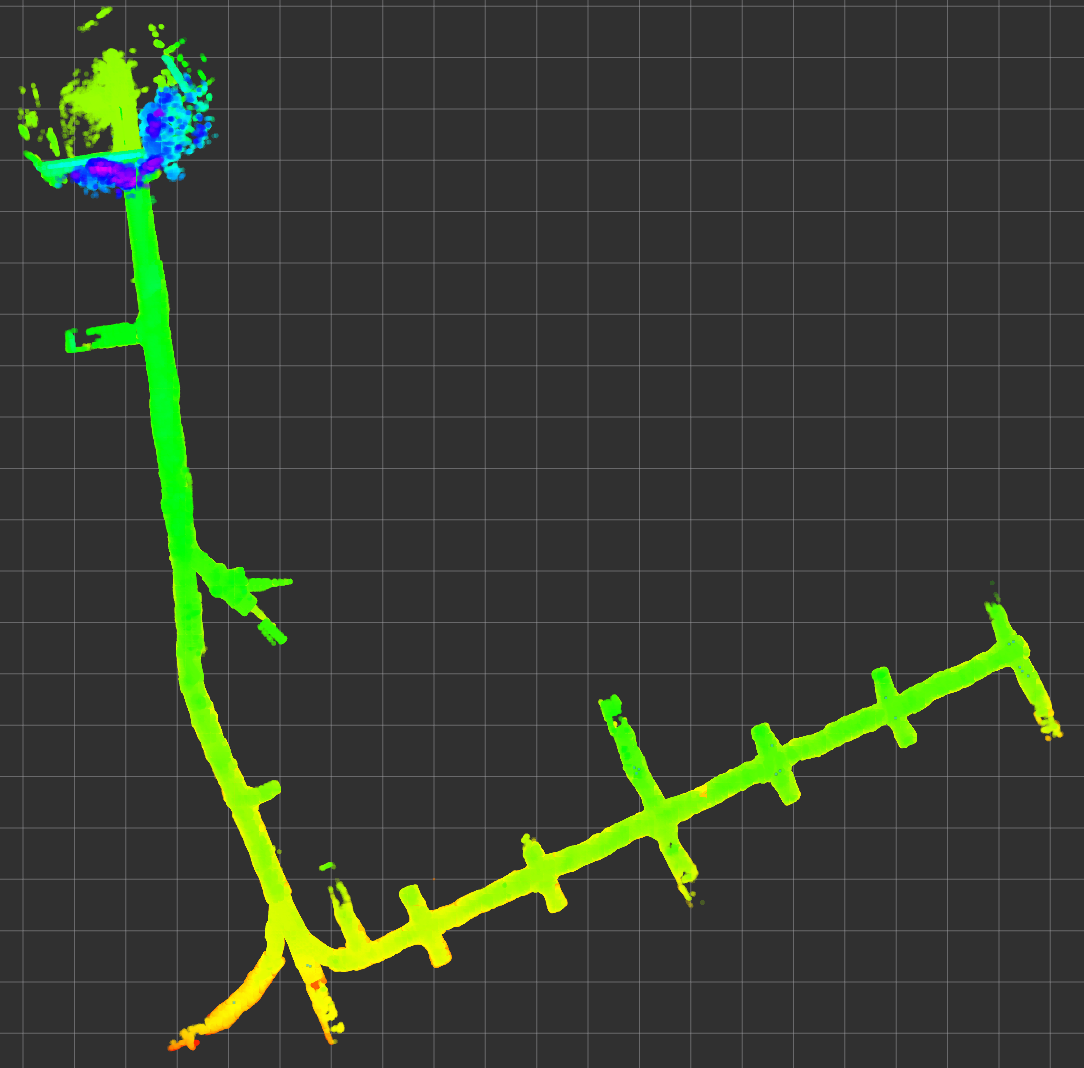}
    \caption{Top view of a Josef mine from a manually driven dataset. At the time, we have focused on planar mapping only. 10m grid.} 
\end{center}
\end{figure}
\rnd{
Additionally, we encountered issues with using RGBD cameras that could not reliably be used in dark places with CCTV surveillance cameras since infrared light from them interfered with the projected patterns of the camera.}
\subsection{Tunnel}

In total third but first contest deployment in the competition took place as a part of the DARPA SubT ``Tunnel'' circuit.
This setting was divided into two separate tracks called `Safety research mine' (SR) and `Experimental mine' (EX), which both took place in the coal mine of the National Institute of Occupational Safety and Health.
(NIOSH).
Each track had two runs, each lasting one hour, \rnd{during which all artifacts that robots had in the line of sight were detected and scored.}
\subsubsection{Run 1, SR}
\begin{figure}[!htb]
\begin{center}
    \includegraphics[width=1.0\columnwidth]{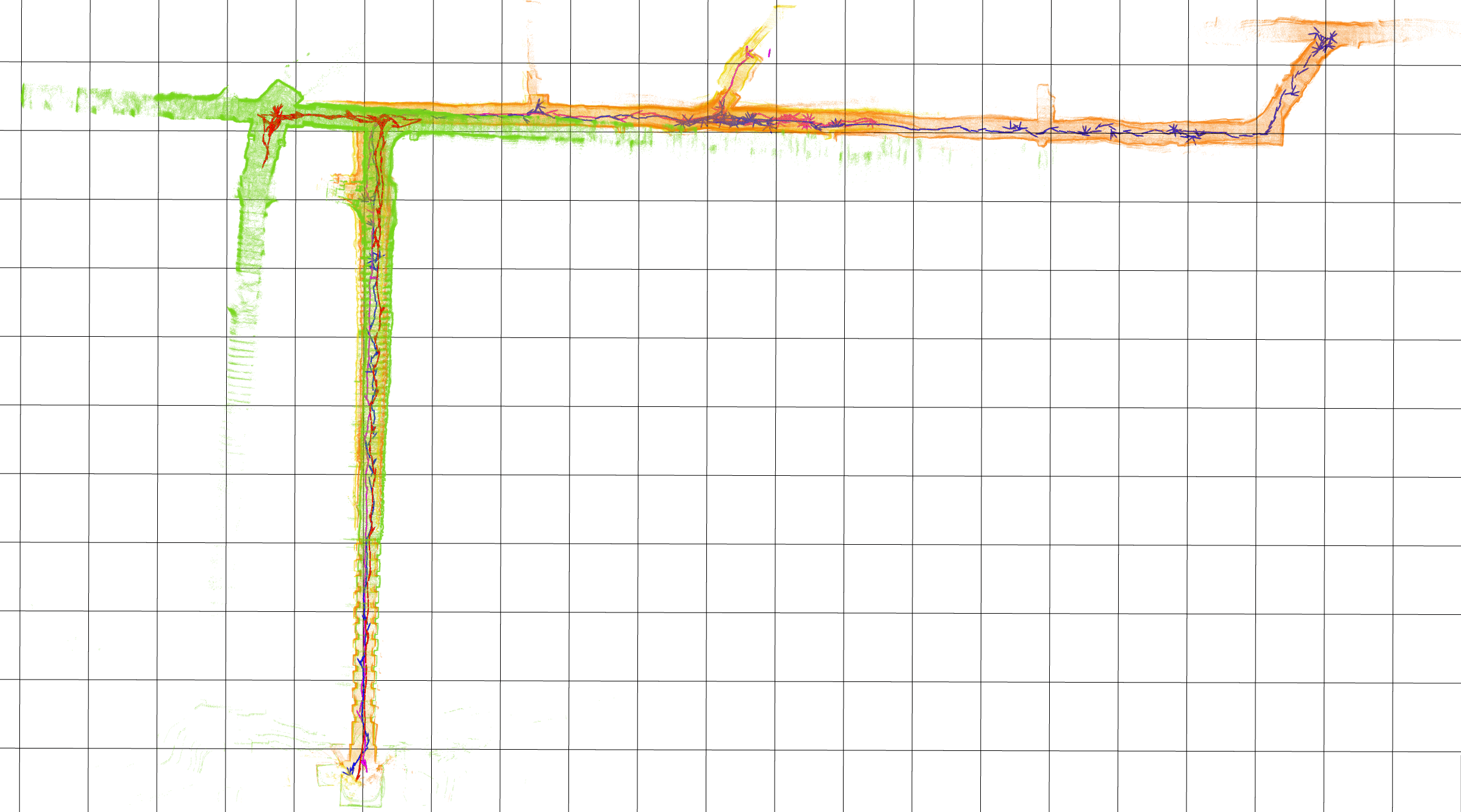}
     \includegraphics[width=1.0\columnwidth]{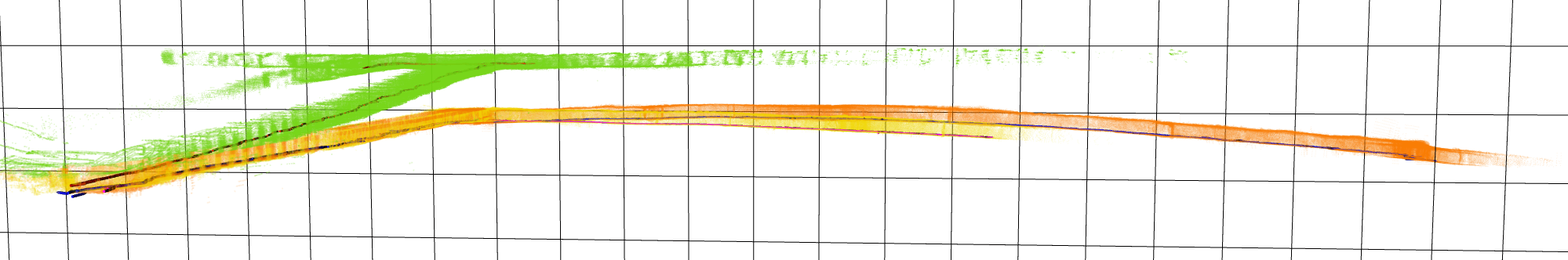}
    \caption{\rnd{Top and side view of a Run 4 in Tunnel circuit. Slight misaligment of robots at the beggining causes visible error in mapping clearly visible at T juntion. Bottom image shows heavy drift in localisation and map bending. 10m grid}}
    	\label{fig:tunnel-bending}
\end{center}
\end{figure}

The first run ended with only one artifact \rnd{(backpack)} found due to miscommunication between the DARPAs emergency stop system and our robot's receiver, which was wrongly configured.
This meant that robots could not venture further than 10 meters into the tunnel due to constantly receiving estop from a DARPA. The operator was able to reprogram one robot on a fly from outside the tunnel with limited bandwidth only.
In the end, we had only eight minutes of mission time for a single robot that was able to find one artifact which was visible from the entrance.\rnd{
In this run, there was no autonomy involved on UGVs all drive was done by remote joystick. However, autonomous UAVs were deployed last minute to test the flight capabilities.
One UAV could go about 100m into the tunnel, and while heading back, it hit a wall with its propellers due to low ceiling and crash-landed.
The other UAV was tasked with going as deep into the mine as possible and having managed to travel 200m into the tunnel, where it stopped due to the limited map size.}
The emergency stop issue was fixed for the other three runs.
\rnd{ 
  \subsubsection{Run 2, EX}
During the second run were able to deploy all robots into the tunnel.
The tracked robot reached 190m into the straight tunnel, where it detected a cellphone.
This artifact has cost us ten tries to report due to poor robot localization drift, which exhibits itself by `map bending', see in Fig.\ref{fig:tunnel-bending}.
The mislocalization had to be corrected manually by the operator by moving the estimated detection upwards in the GUI. 
The error was a 10m drop on the 200m distance driven, which was previously unobserved.
Autonomy, where the operator is supposed to only set waypoints had all robots driving slowly due to the width of the tunnel where robots tended to 'look' from side to side, causing a zigzag trajectory.
This side-to-side movement and constant turning inevitably caused two robots to either crash into a wall or have their motors die due to high loads.
The operator tried to mitigate these problems in at least one robot by driving it using ``force fallow''(FF) markers which the robot follows regardless of obstacles by a straight line.
}

\rnd{
UAVs were again deployed in this scenario, but both were required to return back to the base since they had no wireless link back to the base implemented.
One of the UAVs crashed about 60m in while the other went 100m in and then 80m back, landing out of the crew's reach to gather any data.
For the subsequent runs, the drones were connected to the WiFi network to send data back to the base if they are close enough to the course entrance.
}

\rnd{
Due to the map bending on tracked robots, we have recalibrated the localization stack and forced it not to use the gravitational vector from the IMU.
The husky platform also showed issues with localization robustness since it did not have enough computational power to run ICP at a sufficient frequency.
This was corrected by implementing the T265 Realsense optical odometry camera, which was subsequently mounted with additional lights on the front of the robot to utilize the ceiling of the tunnel for localization to help ICP convergence.
Additionally, threshold values for traversability were tweaked, so the robots were not driving zigzag patterns through the tunnel.
}
\rnd{
\subsubsection{Run 3, EX}
The third run began with a husky platform driving about 80m in a few minutes, with only a suggestion of direction from the operator, where it had one of its power fuses overloaded and shut down.
This effectively rendered the whole platform immovable in the middle of the tunnel and shut down its communication links and retranslating capabilities.
Other tracked robots had issues with traversability due to last-second tweaks, which caused all tracked robots to drive into the walls, where they often got stuck.
This again forced the operator to babysit the robots and use operator GUI directly via ``force fallow''.
One of the robots was sent by the operator to the side tunnel, where it lost a signal and then stopped there since it could not receive new commands. 
}

\rnd{
UAVs in this run had issues with a start where they took of and hit the ceiling just after entering the course.
Both of those flights tripped safety on the drones where they landed back to the ground safely.
This issue happened due to the misconfiguration of the flight level.
}

\rnd{
Since this run, all our communication infrastructure that is supposed to retranslate any data was powered by its own small lithium battery to avoid issues with robots shutting down.
}

\rnd{
\subsubsection{Run 4, SR}
During the last run, no critical hardware issues occurred, making it possible for the robots to traverse more distance than in all other runs combined.
The husky platform had an issue with frontier exploration where few of them stayed outside of the course around the base where the Husky wanted to navigate constantly.
This had to be manually mitigated by controlling the robot directly by waypoints which meant that the robot was driven only to the first crossroads, where it stayed as a retranslation unit for the rest until others could not be driven further.
In the end, the platform was waypoint driven into the unknown by the operator until the signal was lost. Afterward, the robot switched to frontier exploration, which effectively meant that it tried to
return to base.
The operator had to fight this issue by periodically sending new waypoints.
}

\rnd{
Tracked robots were driven by the operator by providing them with waypoints. 
One of the tracked robots got stuck in a small, water body that it could not sense by any means, making it not traversable terrain.
The second was driven 250m into the mine, where it began autonomous frontier exploration two minutes before the end of the run. 
The operator placed the third along the way to retranslate data back to base.
}

\rnd{
  Throughout the whole tunnel circuit, the fully autonomous behavior (frontier exploration) was rather a hindrance since it was fighting the operators' decisions rather than helping him.
  About 90\% of all driving was done by the operator using assistive commands ``force fallow'' and waypoint driving which required him to have a vague sense of direction where the robots can be sent.  
Due to scoring all artifacts that were visible, our team tried to focus more on the reliability of movement and autonomy for subsequent rounds.
}

\rnd{
Due to issues with aligning the internal robot maps with the DARPA frame, we have introduced a total station that is able to localize the robot in a global frame with sub-millimeter precision.
This also allowed us to gather better ground truth data which was previously impossible to do since we did not own any equipment of precise long-range measuring.
}
Preparing for the Urban circuit, we were allowed to perform tests in undisclosed sections of the Prague subway.
This helped obtain a deeper insight into radio communications issues in urban structures and on the traversability of narrow passages, bridges, and stairs.

\subsection{Urban}

So far, the last official deployment was performed during the ``Urban'' circuit in ``Satsop'', which is an unfinished power plant.
Both tracks of this circuit were located around the power plant's unfinished reactor.
\rnd{
In total, our robots drove about 2.5Km and flew 500m, while all artifacts that robots had in the line of sight were detected and scored. }
\subsubsection{Run 1, Beta}
During the first run, most ground robots had severe mobility problems because of the debris scattered on the ground.
\rnd{Husky platform wrongly assumed drivable terrain over a laying beam on which it gets stuck by balancing on its undercarriage.
Husky got freed by another tracked robot that bumps into the stranded robot, making it tip over and eventually break free.
Husky then explores the unknown until it loses link to the base and explores autonomously 20 meters in after it returns to the signal. 
The Husky `rescue' caused the other tracked robot to lose mapping and got lost.
This was caused by the pivoting lidar error, which stopped moving and therefore not returning 3D point clouds necessary for localization.
The second robot got stuck in between two concrete blocks, out of which he could not get.
The last tracked robot was wrongfully planning over uneven terrain even tho it could be driven around.
The robot had issues with an exploration algorithm that was often planning back to the base after it was commanded by the operator to travel deeper into the course.
}

\rnd{
During this run, both drones crashed, entering the course to the door frames.
The liftoff also caused the dust to be perturbed, making it impossible for the small hexapods to navigate.
}

\subsubsection{Run 2, Alpha}
The second run had a significant problem with the localization of one of the tracked robots.
\rnd{This again happened due to the robot crashing with its pivoting Lidar to a wall which it deemed traversable, making it stop.
The operator took full responsibility for this event since he tried to drive the robot using a joystick and the cameras manually.
Both of those data streams had a lag of few seconds, and the operator GUI was constantly crashing while this happened.
}

\rnd{
Two tracked platforms were able to traverse stairs to a lower level of the track.
This traversal was done manually for the first robot with the help of a camera and joystick with a good data stream since the robot was still in range of high bandwidth network to the base.
The second tracked robot first autonomously drove over our small hexapod platform, which acted as a Mobilicom retranslation unit.
Then the robot waited on each mezzanine of the stairwell for approval from the operator to proceed.
Otherwise, the descent was autonomous.
Both downstairs robots were driven by the operator using waypoints in the GUI.
One lost a localization due to a mapping error, whereas the other was autonomously driving back to base if the operator did not set it a waypoint in the last 30 seconds. 
This is especially difficult since the communications allowed the operator to command the robot with about 10 seconds delay.
The robot kept returning to the signal because it had an unexplored frontier back at the base.
}

\rnd{
Husky platform explored the top floor around the reactor where it once got into a position it could not get out of due to perceived obstacles in its surroundings.
The operator's input freed it by issuing the ``force fallow'' command tough GUI two minutes before the end of the run.
}

\rnd{
The UAVs both entered the course without any issue where one crashed after about 50 meters to a wall due to a glitch in localization.
The other traveled about 100m into the course and ended hovering above one place since it could not navigate back to the base.
It safely landed before the batteries run out.
}

\rnd{
Between runs 2 and 3, several days worth of works was done on robots, such as improving the stability of GUI, implementing a reset function for mapping based on the last failed map, ability to restart any
part of the robot for the operator and health status of mapping.
Additionally, more problems were discovered in the pipeline, such as a wrong time setting of the robots when they could not connect to the internet after boot.
This caused WiFi detection to be unusable since they relied on time synchronization between all onboard computers of robots.
}

\subsubsection{Run 3, Beta}
\rnd{
For this run, the traversability of the robots was lowered so they would not drive over debris that was dangerous for them to get stuck on.
This caused other issues, such as almost no robots wanted to traverse anything that was not totally flat; therefore, the robots had to be driven using ``force fallow'' to a waypoint in an operator GUI.
The majority of this run had to be supervised by the operator in this way.
}

\rnd{
Pivoting Lidar of one tracked robot had a hardware failure that caused it to lose localization.
After several remote restarts, it was deemed unsavable and was manually driven using a joystick by the operator to the crossroads to serve as a retransmission unit.
}

\rnd{
Another tracked robot tried to climb stairs to the upper level on the operators' command.
During the climb, the motor controllers reset due to overload, which was commonly encountered on this aging hardware.
In such a situation robot runs the preprogrammed policy from the tunnel circuit. It begins to restart all software and then hardware by importance.
This only causes a robot to stop, but if the actual motor controllers have to be restarted, the flippers can not support the robot's weight.
Since this happened on the stairs, the robot relied on the back flippers to hold it in place, the flippers loosened, and the robot tumbled down the stairs.
No parts except 3D printed mount for depth cameras were broken, and the robot continued to function albeit not movable.
The operator shut down the robot remotely since it was turning its flippers indiscriminately due to a pose it has never observed (undefined behavior).
}

\rnd{
The last tracked robot followed the same path as in Run 1.
Drones again did not make it through doors.
During this run, a brief power outage caused our operator to lose communications to most of the robots and a display of half the base station.
The deployed system had no issue since it primarily relied on battery-powered infrastructure, including our networking.
\begin{figure}[!htb]
\begin{center}
    \includegraphics[width=1.0\columnwidth]{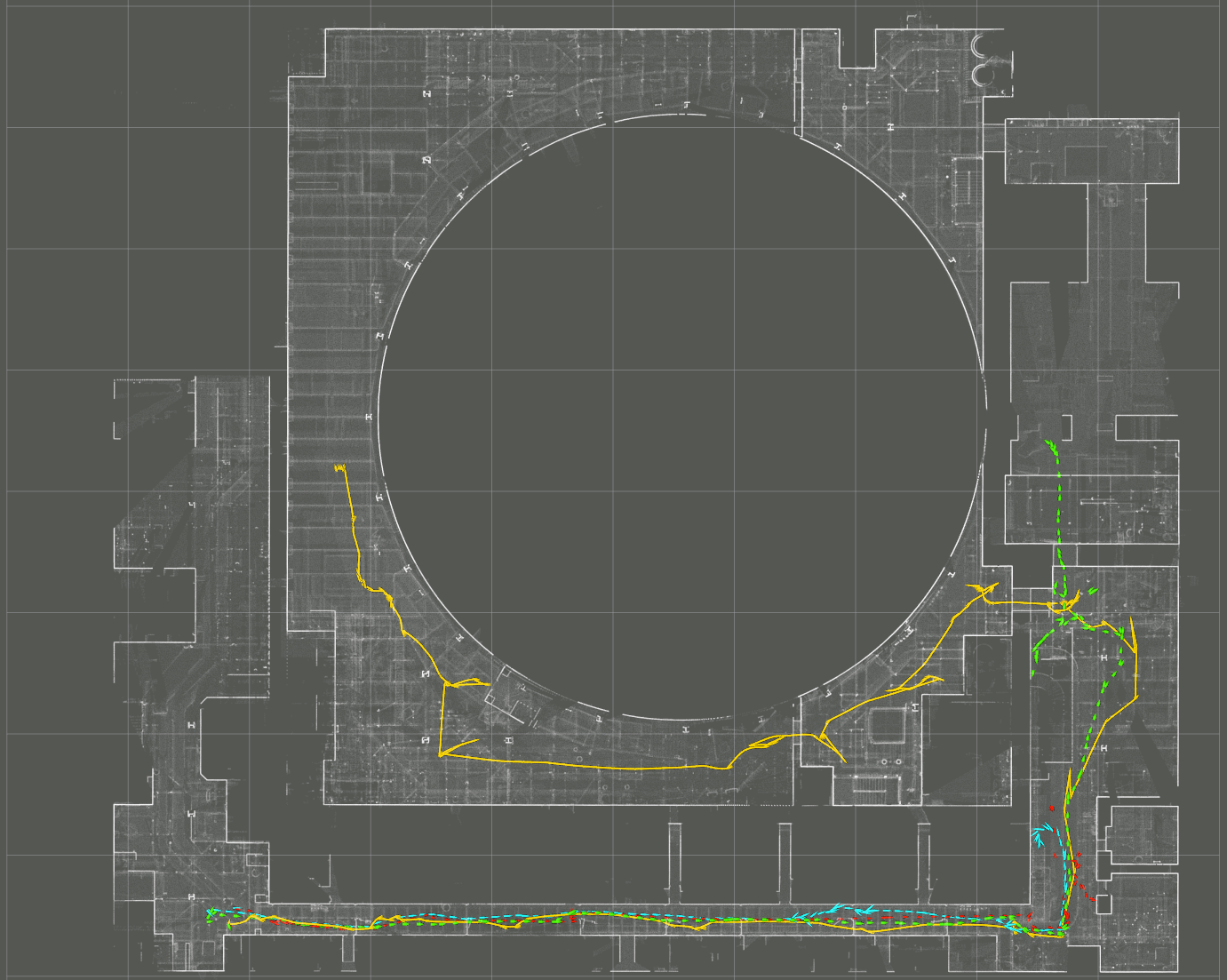}
    \caption{\rnb{Ground truth map of the Beta course in Urban circuit. Only top floor since our team was unable to explore any other level.  Robot positions overlayed in colors each representing a different robot during a run 3. 10m grid.}}
     \label{fig:run3}
\end{center}
\end{figure}
}

\subsubsection{Run 4, Alpha}
\rnd{
In this run, one tracked platform went down the stairs fully autonomously, with only a confirmation from the operator that it can be done.
It autonomously explores about 150m of the lower level where snapped off all of its droppable communication units due to a turn that was too close to an obstacle.
This was caused by not adding the communication units to the robot model, which were protruding out to the back of the robot.
The unintended drop did not impact the performance of the network at all since this communication interface was in range all the time.
}

\rnd{
Husky and other platform explore the top floor, where the tracked platform is mostly command driven by the operator using waypoints since he was afraid that it might get stuck on something with its
problematic motor controllers.
Husky drove the whole path in autonomous exploration mode.
The last tracked robot was not localized in time by the total station in the base, so an impromptu connection to the robot from the total station to the robot had to be made via wire.
Due to this error, we have positioned the robot into the track via joystick about 40 minutes into the run with improper localization to the DARPA frame to act as a retranslation unit for other robots.
}

\rnd{
During this run, two drones were sent in.
One was programmed to fly close to the reactor and then land with a Mobilicom unit. This happened flawlessly.
}
The other UAV crashed and burned down during autonomous exploration.
\rnd{
The reason for the crash was a collision with a beam that forced the drone to drop fast, braking its plastic landing feet.
Due to a rough landing, battery lead insulation was damaged, and the battery shorted itself, subsequently catching on fire.
The fire destroyed most of the platform, including all sensors except Ouster lidar.
Data was later recovered from the onboard computer drive.
}
Even though all those mishaps, we managed to secure third place overall while being first as a self-funded team, i.e., the ranking was the same as in the ``Tunnel'' circuit.

\begin{table}[!ht]
\begin{center}
\caption{\rna{Scored artifacts during the urban round of the competition. Any score point that has been detected by multiple robots is indicated by the brackets.  False positives (FP) are combined
    from each robot separately, meaning one falsely detected object from all robots would account for four FP.}}
\label{tbl:urban_scoring}
\begin{tabular}{c|ccccc|cc}
	Run & Backpack & Vent & Survivor & Cellphone & Co2 & Total& FP\\
	\hline
	1 & 1 & 1 & 0 & 0 & 1 & 3 & 31 \\
  2 & 2 & 2 & 0 & 0 & 0 & 4 &71\\
  3 & 0 & 1 & 1 & 1 & 1 & 4 &55\\ 
  4 & 1 & 3 & 1 & 0 & 1 & 6 & 54\\ 
	
\end{tabular}
\end{center}
\end{table}

\begin{figure}[!htb]
\begin{center}
    \includegraphics[width=1.0\columnwidth]{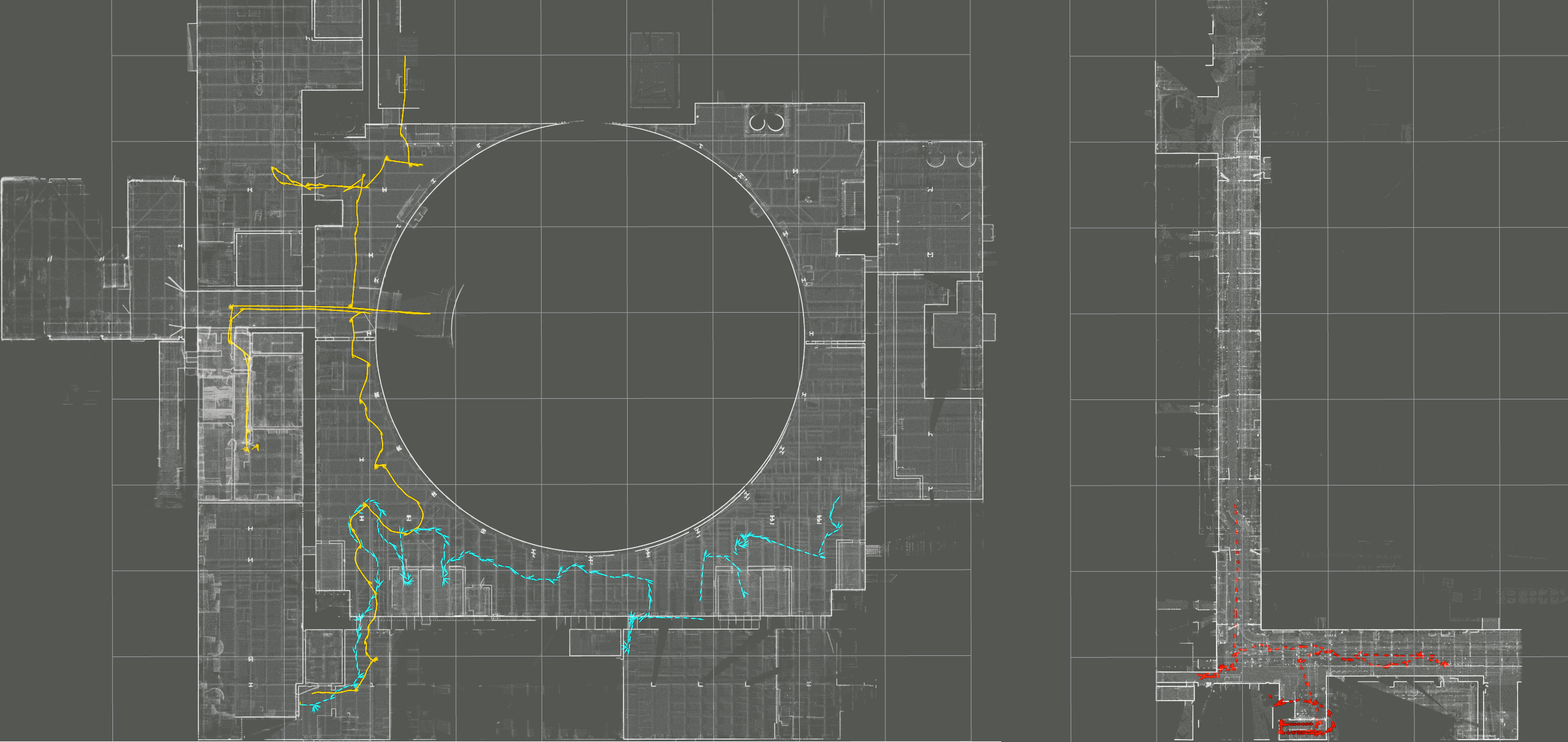}
    \caption{\rnb{Ground truth map of the Alpha course in Urban circuit with upper (left) and lower (right) floors. Robot positions overlayed in colors each representing a different robot during a run 4. 10m grid.}}
    \label{fig:run4}
\end{center}
\end{figure}

\rnd{
\subsection{Bull rock}
In anticipation of the cave circuit, our team went testing to a ``Byci skala''  (Bull rock) natural cave located in Moravia.
This environment is somewhat modified for easier walking having a pathway that comprises parts of mud, sand, rock, and concrete bridges without a railing.
The cave is about 250m in length with steep inclines and declines, which are on the main path eased up by tall stairs sculpted out of the mud.
}

\rnd{
Multiple runs with all robots were made with or without autonomy to find answers to several questions.
Firstly, we tested the capability of tracked robots on stairs without distinct ridges made out of mud, which was difficult to traverse due to slippage.
Additionally, we have tested autonomous navigation with negative obstacles, which our team did not have a chance to test before except stairs.
This led to the flipping of a Husky robot that ran off to a side caused by elevation mapping where negative obstacles were not inflated as walls were.
All parts of the Husky robot stayed intact, including the lidar sensor, which held the whole weight of the robot without any roll cage.
First tests with a ballon-like device were tested mostly for abrasion against the low ceiling of the caves. Latex-based ballons proved to be useless since they lost most of their helium through porosity. 
Foil-based balloons showed promises of being able to withstand pulling over rough rocks. 
}

\rnd{
During our trials, we were able to traverse most of the cave autonomously, albeit with close supervision, except the stairs part, which has proven to be difficult for all UGVs.
Tracked robots had issues with battery life to a level requiring two or three batteries spare to travel in and out of the cave.
}

\rnd{
Tests of detecting water for avoidance were conducted using a Husky platform for driving in a shallow (5cm) water stream that runs throughout the cave.
All attempted solutions to this, including usage of stereo cameras, polarized light, automotive radars, and sonars, did not yield any acceptable results.
}

\rnc{
Additionally, comms propagation tests showed that in the low branching environment we had, it is possible to interconnect the whole cave using only four units of Mobilicom devices if placed optimally.
Complex tests with MOTE units were also conducted to improve com drop predictions as show in~\cite{zoula2021building}.
}

\rnd{
During our experiments in the cave, some robots have suffered from problems with condensation.
The cave has a temperature of about eight °C and nearly 100\% humidity. This causes all robots to be cooled down to a temperature that can cause condensation on platforms while leaving the cave. 
We eventually had an issue with a Husky platform not working, most likely due to water condensed on internal components. 
}

\rnd{
\subsection{Virtual}
Due to the cancellation of the Systems Cave circuit, we could only participate in the Virtual part of the challenge with which we had no prior experience.
Before we have decided to participate, the only experience we had with simulation was from drones and traversability experiments with Absolem, which made it quite challenging for us to establish a workflow.
To make it more complicated, our private computational grid only supports Singularity HPC containers, while the SubT challenge only supports Docker containers.
Another significant difference of the competitive environment was the use of Ignition Gazebo, as all of our previous simulations were running in Gazebo Classic, which is very different from Ignition Gazebo.
}

\rnd{
\subsubsection{Cloudsim}
SubT Virtual challenge offers the Cloudsim web simulation environment to registered teams. It can run up to three simulations (packaged as Docker images) simultaneously, each with a real-time factor $3 \%$ -- this means that each 1-hour simulated run takes about 30-40 hours of real-time.
The teams do not have access to the internal state of the simulation during its run.
All interesting data for the teams have to be packed in a 2~GiB ROS Bag file per simulated robot (this especially means it is impossible to store all camera images -- not even compressed).
We used DRACO compression for pointclouds which at least allowed us to store all pointclouds collected during the simulation.
When the simulation is finished, the teams are given the ROSBag files together with the ground truth positions of the robots (which are not accessible during the simulation).
}

\rnd{
\subsubsection{Porting Docker images to Singularity}
The inability to observe the internal state of the simulation and custom code when it is running is very limiting in debugging.
Put together with the fact that each simulation runs 30-40 hours and cannot be preempted, using only Cloudsim for debugging would be almost impossible.
}

\rnd{
We thus put effort into transforming the provided Docker images into Singularity containers. Singularity offers a tool to import Docker images, and it works well. However, there are differences in the philosophy of Singularity compared to Docker, which required further effort.
}

\rnd{
Most importantly, Singularity mounts the home folder of the user running the container and makes it accessible in the virtual environment, together with most environment variables. So wherever the Docker image assumes that \textit{HOME} environment variable will always be \textit{/home/developer}, it is different in Singularity.
Moreover, the SubT virtual simulator hardcodes the \textit{/home/developer} path at some places of its code.
These places need to be manually found and substituted with the \textit{HOME} variable -- but now always.
The \textit{/home/developer} path is used in two different contexts: 1) to point to pre-built simulator files, which are stored under \textit{/home/developer} when building the Docker container, 2) as a relative location of configuration and log locations used by ROS and Ignition Gazebo (e.g. \textit{~/.ros/log} for ROS logs).
All occurrences of context 1 have to be replaced with \textit{/home/developer}, and all occurrences of context 2 have to be replaced with reference to \textit{HOME} variable (to store the logs in the host user's home directory).
}

\rnd{
Another significant difference is that the Singularity images and containers are immutable.
So the only two writable locations are, by default, the user's home directory and \textit{/tmp} folder (which is also shared with the host system).
This is why writing logs and other generated data in \textit{/home/developer} is not possible.
}

\rnd{
The last major difference is that the user has the same permissions towards the rest of the system inside the Singularity container as he or she has in the host system.
So if the user is not allowed to call \textit{sudo} in the host system, calling it inside the container is also not allowed (not even during the build of the container).
This is why we have to build the images on personal computers (where we have root privileges) and upload the built images to the computation grid (where we only have standard user privileges).
}

\rnd{
With all these differences resolved, we can build Singularity images and run them on the local computation grid.
When the simulation is running in this grid, we have full access to both the simulator and custom code internal state, we can observe topics, and we can even interact with the simulation, pause it, move the robots, view the raw images from cameras and so on.
This dramatically increases the efficiency of debugging problems with the simulation.
}

\rnd{
\subsubsection{Differences between Real and Simulated Missions}
We were looking forward to working with a fleet of robots that do not break or fall apart after every mission in a hardware sense.
However, it showed that even the simulated robots have their problems that need to be addressed.
}

\rnd{
A big difference was the simulation of tracked robots.
Ignition Gazebo, unfortunately, can not simulate tracks in a faithful way, so the main tracks of Absolem were simulated by eight wheels.
This results in very different behavior of the robots in complex terrain, and usually, the wheels have worse traversability -- it is, e.g., almost impossible to climb up staircases with the wheel-based tracks.
This made the Absolem robots much less useful in the simulation.
}

\rnd{
On the other hand, the Husky robot models provided in simulation worked better than in reality, and as the robot cannot break any expensive sensor, we were able to run it much faster than we dared in the Systems track.
Each Husky robot was thus able to explore up to 3~km of caves in a~simulated run.
Our record from Tunnel and Urban circuits was about 400 meters with Husky.
}

\rnd{
\subsubsection{Path Planning Internal Competition}
As the use of a simulator allows us to do much more tests than what can be done with real robots, we took the opportunity and developed two separate path planners based on different principles.
}

\rnd{
The first one was the \textit{RDS planner},
an evolution of the path planner used in Tunnel and Urban circuits (described in Section~\ref{sec:navigation}).
}

\rnd{
The second path planner, called \textit{Naex}, used an unstructured point-cloud map incrementally built from available RGBD and lidar depth measurements as the primary representation.
The map is accompanied by an approximate nearest-neighbor graph constructed using the FLANN library~\cite{Muja-2009-VISAPP}.
Each point in the map is assigned a reward based on its closest distance to any robot so far and the number of frontier points nearby.
Dijkstra's algorithm, namely the implementation from the Boost Graph Library (see~\cite{Siek-2001-BoostGraph}), is then used to plan the shortest paths to all reachable points, and the point with the highest ratio of reward to distance is selected as navigation goal.
}

\rnd{
As we did not know which path planner would perform better in simulation, we set up an internal competition of the path planners.
We ran dozens of simulations with the whole fleet of robots with each of the path planners.
When the simulations were finished, we evaluated the number of points scored, explored area and other metrics, and chose the Naex planner for use in the scored runs.
}

\rnd{
We have even considered randomizing which path planner would be used, but the idea was rejected in the end as it was not possible to change the robot roster for the scored runs, and Naex worked better with Husky, whereas RDS planner worked better with Absolem.
}

\rnd{
\subsubsection{Deployment}
Converting our system to be deployable as a Singularity and Docker image helped to straighten our workspace build process and catching many minor deployment errors like missing declarations of dependencies, undefined build order, missing install directives, and so on.
}

\rnd{
One of the tricky parts to handle was the detector weights.
They come as a large binary file (about 500 MiB) and change often, so it is not possible to store them as a part of a Git repository as the rest of the system.
So we have them stored in a shared network location via HTTP protocol and add them to the images in a special build step. However, this deployment process seemed to have problems with local caching of the detector weights, so the UGVs were using some older version of the weights in the competition, which hindered their detection capabilities.
}

\rnd{
\subsubsection{Cave Virtual Competition Results}
The scored runs of the virtual competition were run in 8~different virtual worlds, and there were 3~independent runs of the system for each world.
This totals 24 scored runs.
The ranking of teams was done by summing up the number of artifacts their robots correctly reported.
}

\begin{figure}[!htb]
	\begin{center}
		\includegraphics[width=1.0\columnwidth]{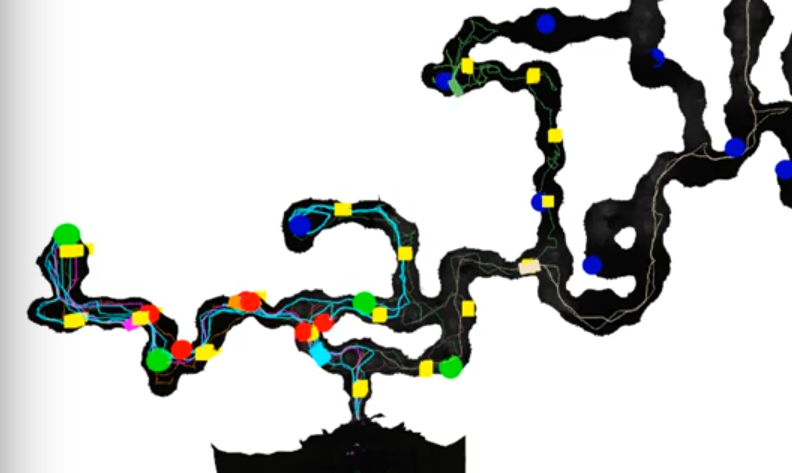}
		\caption{\rnb{\textbf{Path trace of a~scored competition run.} Colored lines are traces of individual robots. Blue circles are ground truth artifact positions. Green circles are scored artifacts. Red circles are unsuccessful artifact reports (wrong position or class). Yellow rectangles are dropped communication motes.} 		}
		\label{fig:path-trace}
	\end{center}
\end{figure}

\rnd{
When data from the scored runs were made available to use, we ran numerous analyses to learn about our system's behavior.
This way, we found the outdated detector weights on UGVs.
An example graphical evaluation of a~run is shown in Figure~\ref{fig:path-trace}.
}

\rnd{
We have composed a table of points scored by each robot.
There is not a~unique link between a~scored point and a~particular robot, as each artifact can be seen and reported by multiple robots.
In this case, we have assigned the point to the robot that saw the artifact first.
The overall scored points are in Table~\ref{fig:virtual-results-table}.
}

\rnd{
Table~\ref{fig:virtual-results-hypotheses} shows how many artifacts were detected and localized at all by each robot.
Some of the artifacts were not reported because the robots did not always return to broadcast their positions in time.
}

\begin{table}[!htb]
\begin{center}
\caption{\rnb{\textbf{Scored points of our robots in scored runs.} Robots \textit{uav2} and \textit{uav4} correspond to models \textit{SSCI\_X4\_SENSOR\_CONFIG\_1}, \textit{uav5} is \textit{EXPLORER\_DS1\_SENSOR\_CONFIG\_1}, and \textit{X1, X2} and \textit{X3} are \textit{EXPLORER\_X1\_SENSOR\_CONFIG\_2} Husky-like ground robots. Run name is composed of numeric ID of the world and index of the run (1-3).}}
\label{fig:virtual-results-table}
\begin{tabular}{c|cccccc|c}
	World-Index & uav2 & uav4 & uav5 & X2 & X3 & X1 & Total \\
	\hline
	1-1 & 1 & 0 & 2 & 1 & 0 & 0 & 4 \\
	1-2 & 0 & 0 & 3 & 0 & 0 & 0 & 3 \\
	1-3 & 0 & 0 & 2 & 0 & 0 & 0 & 2 \\ 
	\hline
	2-1 & 0 & 1 & 3 & 0 & 0 & 0 & 4  \\ 
	2-2 & 1 & 1 & 2 & 0 & 1 & 0 & 5  \\ 
	2-3 & 1 & 2 & 1 & 0 & 0 & 0 & 4  \\ 
	\hline
	3-1 & 1 & 1 & 1 & 0 & 0 & 0 & 3  \\ 
	3-2 & 3 & 0 & 0 & 0 & 0 & 0 & 3  \\ 
	3-3 & 2 & 2 & 1 & 0 & 0 & 0 & 5  \\ 
	\hline
	4-1 & 0 & 3 & 0 & 0 & 0 & 0 & 3  \\ 
	4-2 & 0 & 2 & 0 & 0 & 0 & 0 & 2  \\ 
	4-3 & 1 & 1 & 2 & 0 & 0 & 0 & 4  \\ 
	\hline
	5-1 & 1 & 0 & 1 & 0 & 0 & 0 & 2  \\ 
	5-2 & 0 & 0 & 3 & 1 & 0 & 0 & 4  \\ 
	5-3 & 0 & 1 & 1 & 0 & 0 & 0 & 2  \\ 
	\hline
	6-1 & 3 & 4 & 0 & 0 & 0 & 0 & 7  \\ 
	6-2 & 5 & 2 & 1 & 0 & 0 & 0 & 8  \\ 
	6-3 & 1 & 4 & 2 & 0 & 0 & 0 & 7  \\ 
	\hline
	7-1 & 3 & 0 & 1 & 0 & 0 & 0 & 4  \\ 
	7-2 & 2 & 0 & 0 & 0 & 0 & 0 & 2  \\ 
	7-3 & 1 & 0 & 0 & 0 & 0 & 0 & 1  \\ 
	\hline
	8-1 & 0 & 1 & 1 & 0 & 0 & 0 & 2  \\ 
	8-2 & 0 & 2 & 3 & 1 & 0 & 0 & 6  \\ 
	8-3 & 1 & 2 & 1 & 0 & 0 & 0 & 4  \\ 
	\hline
	Total & 27 & 29 & 31 & 3 & 1 & 0 & 91  
\end{tabular}
\end{center}
\end{table}

\begin{table}[!htb]
\begin{center}
	\caption{\rnb{\textbf{Correct hypotheses of artifact locations.} This table shows the number of correctly placed artifact location hypotheses each robot created during a scored run. Some of the hypotheses were not reported and scored because the robots did not return to comms until the mission end (e.g., because they flipped over or there was a path planning error).}}
	\label{fig:virtual-results-hypotheses}
	\begin{tabular}{c|cccccc|c}
		World-Index & uav2 & uav4 & uav5 & X2 & X3 & X1 & Total \\
		\hline
		1-1   & 1    & 0    & 2    & 1  & 0  & 0  & 4     \\
		1-2   & 1    & 0    & 3    & 1  & 1  & 0  & 6     \\
		1-3   & 0    & 1    & 3    & 1  & 1  & 1  & 7     \\
		\hline
		2-1   & 2    & 1    & 3    & 0  & 3  & 3  & 12    \\
		2-2   & 2    & 1    & 3    & 2  & 1  & 0  & 9     \\
		2-3   & 2    & 3    & 2    & 1  & 1  & 3  & 12    \\
		\hline
		3-1   & 3    & 1    & 3    & 2  & 3  & 1  & 13    \\
		3-2   & 4    & 2    & 2    & 1  & 2  & 1  & 12    \\
		3-3   & 3    & 3    & 1    & 2  & 2  & 2  & 13    \\
		\hline
		4-1   & 0    & 3    & 2    & 0  & 0  & 0  & 5     \\
		4-2   & 0    & 2    & 2    & 0  & 0  & 0  & 4     \\
		4-3   & 1    & 1    & 2    & 1  & 0  & 0  & 5     \\
		\hline
		5-1   & 2    & 0    & 2    & 1  & 1  & 3  & 9     \\
		5-2   & 0    & 1    & 4    & 4  & 2  & 7  & 18    \\
		5-3   & 0    & 2    & 1    & 2  & 1  & 6  & 12    \\
		\hline
		6-1   & 3    & 5    & 0    & 3  & 0  & 0  & 11    \\
		6-2   & 5    & 2    & 2    & 0  & 3  & 0  & 12    \\
		6-3   & 2    & 4    & 4    & 0  & 0  & 0  & 10    \\
		\hline
		7-1   & 3    & 0    & 1    & 0  & 2  & 2  & 8     \\
		7-2   & 2    & 0    & 0    & 6  & 1  & 0  & 9     \\
		7-3   & 2    & 0    & 0    & 0  & 0  & 0  & 2     \\
		\hline
		8-1   & 0    & 1    & 1    & 1  & 1  & 1  & 5     \\
		8-2   & 0    & 3    & 4    & 1  & 1  & 1  & 10    \\
		8-3   & 1    & 2    & 3    & 1  & 1  & 1  & 9     \\
		\hline
		Total & 39   & 38   & 50   & 31 & 27 & 32 & 217  
\end{tabular}
\end{center}
\end{table} 

\section{Conclusion}
\rnd{
Team CRAS-CTU-NORLAB has deployed a multi-robotic system in diverse subterranean environments during DARPA SubT.
Our team approach for the contest was primarily focused on single-agent robustness to mitigate possible issues with a small robot fleet.
This resulted in most advances being made in localization, communications, and detections.
Robust ICP based localization allowed for rather rough handling of the robots during the drive, which is typical when small platforms traverse harsh terrains.
Implemented object detection pipeline based on YOLOv3 showed its strengths mainly in a way that no object has been missed.
Most of the runs were either teleoperated or heavily influenced by the operator who was able to control robots via waypoint navigation most of the time due to our multi-network based communications.
}

\rnd{
The first round of the competition called ``Tunnel'' was held in a coal mine where robots searched for artifacts in repetitive tunnels and long hallways with low ceilings and water-soaked ground.
In the second ``Urban'' scenario in February 2020, robots had to operate in the urban underground with service stairs, metal bridges, small rooms, metal fixtures, and holes in the ground.
This circuit took place at an unfinished nuclear power plant.
In both official deployments, our team managed to secure the third rank while being the winner among non-DARPA funded teams~\cite{DARPA2019}~\cite{DARPA2020}.
}

\rnd{
Additionally to the systems competition rounds, our team participated in the Virtual Challange as well after it was announced that the system track is being canceled due to a global pandemic.
After a long struggle, we have managed to secure sixth place.
We consider it a reasonable result given that we had only four months to port our pipelines into the simulated environments we were not familiar with
}

\rnd{
In the future we plan on expanding our knowledge in multi-robot cooperation, a better understanding of the signal propagation as well as the implementation of new platforms such as dog-like robots or large
hexapods.
}

\bibliographystyle{apalike}
\bibliography{main}

\end{document}